\documentclass[10pt,journal,compsoc]{IEEEtran}
\usepackage[colorlinks,linkcolor=red]{hyperref}
\usepackage{multirow}
\usepackage{array}
\usepackage{bbding}
\usepackage{graphicx}  
\usepackage{tablefootnote}
\usepackage{flushend}

\usepackage{tikz}
\usetikzlibrary{shapes.geometric, positioning, arrows.meta}
\usepackage{pgfplots}
\pgfplotsset{compat=1.16}

\ifCLASSOPTIONcompsoc
  \usepackage[nocompress]{cite}
\else
  \usepackage{cite}
\fi

\ifCLASSINFOpdf
\else
\fi

\begin{document}
%
\title{WGSR-Bench: Wargame-based Game-theoretic Strategic Reasoning Benchmark for \\Large Language Models}
%
%
%
%

\author{Qiyue Yin, 
        Pei Xu,
        Qiaozhe Li,
        Shengda Liu,
        Shengqi Shen,
        Tong Wang,
        Yihong Han,
        Xiaonan Zhao,
        Likun Yang,
        Shiyue Cao,
        Shiyu Qiu,
        Yuxuan Liu,
        Shizhao Yu,
        Lei Cui,
        Chengxin Yan,
        Jie Sun,
        Xiangquan Tang,
        Kaiqi Huang 
\IEEEcompsocitemizethanks{\IEEEcompsocthanksitem Qiyue Yin, Pei Xu, Qiaozhe Li and Shengda Liu contribute equally (qyyin@nlpr.ia.ac.cn, (pei.xu, liqiaozhe2015, shengda.liu)@ia.ac.cn).
\IEEEcompsocthanksitem Qiyue Yin, Pei Xu, Qiaozhe Li, Shengda Liu, Shengqi Shen, Tong Wang, Yihong Han, Xiaonan Zhao, Likun Yang, Shiyue Cao, Shiyu Qiu, Yuxuan Liu, Shizhao Yu, Lei Cui, Chengxin Yan, Jie Sun, Xiangquan Tang, Kaiqi Huang are with Institute of Automation, Chinese Academy of Sciences, Beijing, China, 100190. \\
Qiyue Yin and Kaiqi Huang are also with University of Chinese Academy of Sciences, Beijing, China, 100049.
\protect
\IEEEcompsocthanksitem Corresponding author: Kaiqi Huang (kqhuang@nlpr.ia.ac.cn) \protect 

}
}

%
%

\markboth{Journal of \LaTeX\ Class Files}%
{Shell \MakeLowercase{\textit{et al.}}: Bare Demo of IEEEtran.cls for Computer Society Journals}
%



\IEEEtitleabstractindextext{%
\begin{abstract}
Recent breakthroughs in Large Language Models (LLMs) have led to a qualitative leap in artificial intelligence’s performance on reasoning tasks, particularly demonstrating remarkable capabilities in mathematical, symbolic, and commonsense reasoning. However, as a critical component of advanced human cognition, strategic reasoning, i.e., the ability to assess multi-agent behaviors in dynamic environments, formulate action plans, and adapt strategies, has yet to be systematically evaluated or modeled. To address this gap, this paper introduces WGSR-Bench, the first strategy reasoning benchmark for LLMs using wargame as its evaluation environment. Wargame, a quintessential high-complexity strategic scenario, integrates environmental uncertainty, adversarial dynamics, and non-unique strategic choices, making it an effective testbed for assessing LLMs’ capabilities in multi-agent decision-making, intent inference, and counterfactual reasoning. WGSR-Bench designs test samples around three core tasks, i.e., Environmental situation awareness, Opponent risk modeling and Policy generation, which serve as the core S-POE architecture, to systematically assess main abilities of strategic reasoning. Finally, an LLM-based wargame agent is designed to integrate these parts for a comprehensive strategy reasoning assessment. With WGSR-Bench, we hope to assess the strengths and limitations of state-of-the-art LLMs in game-theoretic strategic reasoning and to advance research in large model-driven strategic intelligence.

\end{abstract}

}

\maketitle

\IEEEdisplaynontitleabstractindextext

\IEEEpeerreviewmaketitle


\IEEEraisesectionheading{\section{Introduction}}
\IEEEPARstart{I}{n} recent years, the groundbreaking development of large language models (LLMs) has brought revolutionary transformations to the field of artificial intelligence \cite{LLMSurvey}. Foundational LLMs such as GPT-4 \cite{GPT4} and Deepseek V3 \cite{deepseekV3} have demonstrated exceptional capabilities in text generation, language translation, and multimodal understanding. However, these models show limitations when confronted with complex multi-step thinking and reasoning scenarios. In response, reasoning-oriented LLMs \cite{Rmodels1, Rmodels2} represented by OpenAI's o1 and Deepseek R1 \cite{deepseekR1} have emerged, demonstrating remarkable capabilities in reasoning tasks, particularly in mathematical \cite{math}, symbolic \cite{symbolic}, and commonsense \cite{commonsense} reasoning.

Game-theoretic decision-making serves as a general methodology for addressing strategic choices among multiple participants under intertwined interests and rule constraints. It is ubiquitously present across all facets of the real world, ranging from great-power competition down to commodity pricing. As the cognitive engine of game-theoretic intelligence, strategic reasoning \cite{SR} constitutes the cornerstone for achieving situational awareness in games and complex decision-making. Therefore, systematically evaluating the strategic reasoning capabilities of current large models, particularly large reasoning models, has become an urgent research imperative. Analogous to existing benchmarks for assessing reasoning capabilities in large models \cite{Rmodels3}, such evaluations form the foundation for advancing in-depth research on strategic reasoning.

Current benchmarks for strategic reasoning assessment can be broadly categorized into two types.
The first type employs complex game environments such as StarCraft and Avalon to evaluate large model-based agents. These intricate environments demand robust strategic reasoning capabilities to accomplish long-horizon situational awareness, planning, and strategy generation. For instance, Light et al. proposed AvalonBench \cite{AvalonBench}, a representative benchmark for strategic social deduction, which incorporates rule-based AI bots to assess large model-based agents' decision-making abilities in scenarios requiring deception and reasoning. 
Ma et al. introduced TextStarCraft \cite{TextStarCraft1, TextStarCraft2}, a text-based interface for StarCraft environments, and developed a chain-of-summary reasoning method that enables large models to defeat built-in AI bots at difficulty level 5.

\begin{figure*}
   \begin{center}
   \includegraphics[width=0.95\textwidth]{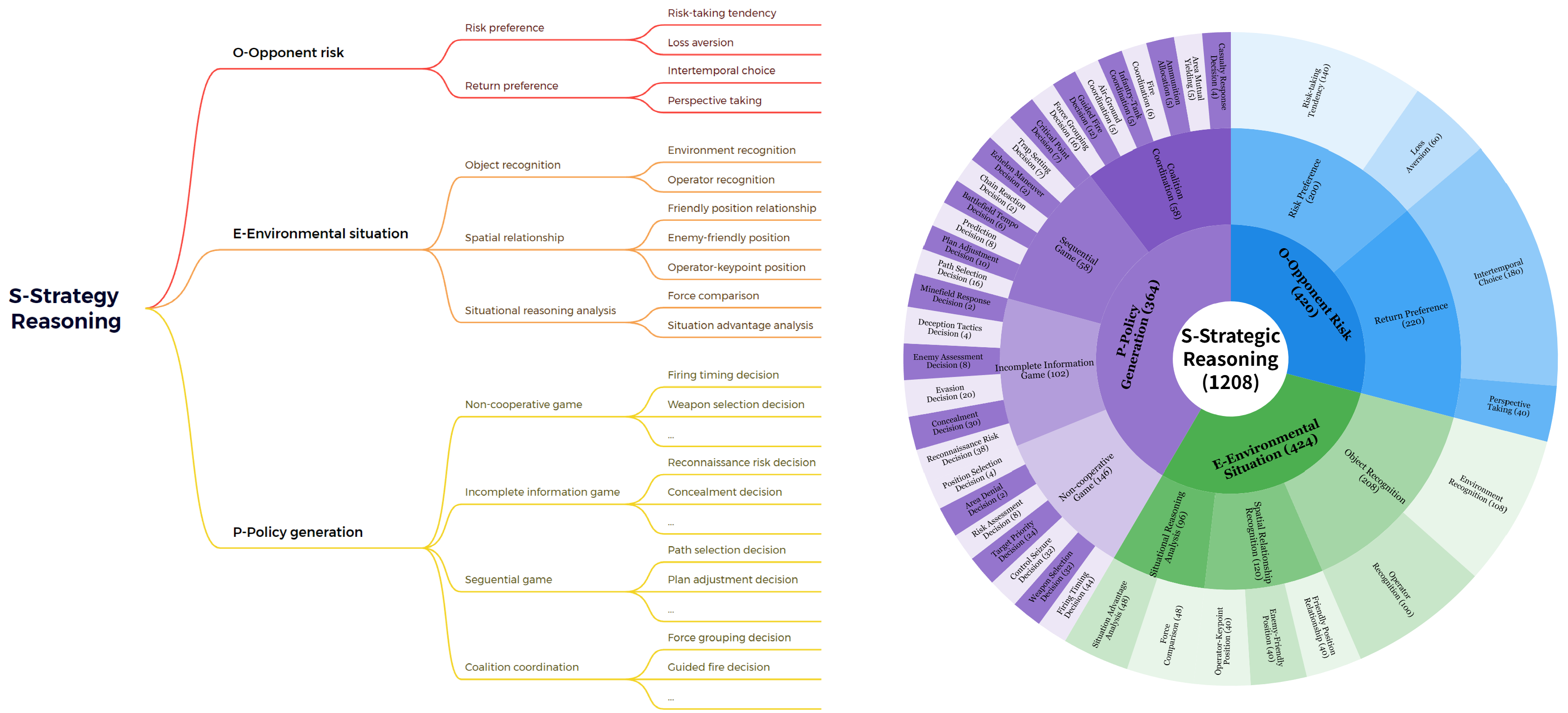}
   \end{center}
   \caption{Overall framework of WGSR-Bench.}
   \label{framework}
\end{figure*}

The second category of benchmarks primarily draws upon classical game-theoretic scenarios to evaluate strategic reasoning capabilities. While these settings are relatively simplified, they inherently possess the capacity to validate the strategic reasoning proficiency of large language models. 
Duan et al. proposed GTBench \cite{GTBench}, a language-driven
environment including 10 widely-recognized tasks for strategic and logical reasoning evalution. 
Askari et al. introduced MAgIC \cite{MAgIC}, which consists of two social deduction games and three game-theory scenarios, to evaluate LLMs' reasoning, planing and other social abilities. 
Wu et al. developed SmartPlay \cite{SmartPlay}, a games collection, to evaluate LLMs as agents, where reasoning, planning, and learning from history abilities are required.

In addition to the aforementioned two categories of strategic reasoning evaluation benchmarks, some researchers have focused on constructing assessment frameworks for key capabilities such as task planning and deception in large language models. These competencies are intrinsically linked to strategic reasoning and have, to some extent, propelled advancements in the study of strategic reasoning in large models. 
Hill et al. introduced MinePlanner \cite{MinePlanner}, which consists of 45 tasks from Minecrarft, for planing evaluation.
Furthermore, Qiao et al. developed WorfBench \cite{WorfBench}, a workflow generation benchmark, which evaluate LLMs' planning abilities, especially for bridging the gap between tasks and specific executable actions.
Wu et al. proposed OpenDeception \cite{OpenDeception}, an open-ended scenario dataset, which systematic assessments LLM-based agents for recognizing deception intention.

In summary, current strategic reasoning evaluation benchmarks exhibit two critical limitations. 
Firstly, existing benchmarks focus solely on end-to-end validation of strategic reasoning capabilities, while strategic reasoning itself constitutes an integration of multiple sub-capacities \cite{SR}, e.g., contextual awareness and opponent empathy analysis. This narrow focus prevents accurate characterization of the proficiency levels in key components of strategic reasoning. 
Secondly, classical game-theoretic scenarios adopted in evaluations are overly simplistic, demonstrating significant deficiencies in reasoning complexity. Conversely, while complex gaming environments like StarCraft provide richer contexts, their evaluation scope remains singular—limited primarily to strategy generation and win/loss assessment, which presents notable constraints in both coverage and comprehensiveness. 
These inherent limitations in current evaluation frameworks substantially impede systematic assessment of strategic reasoning capabilities.

To address these limitations, this paper proposes WGSR-Bench, the first large model strategic reasoning benchmark based on wargame.
As a paradigmatic high-complexity strategic game, wargame \cite{wargame} inherently integrates environmental uncertainty, adversarial dynamics, and non-unique strategic options, thereby providing a rigorous testbed for evaluating model capabilities in multi-agent decision-making, intent inference, and counterfactual reasoning—all critical dimensions of strategic reasoning.
Building upon game-theoretic decision principles, we innovatively develop the S-POE structured cognitive framework for strategic reasoning. 
Here, POE represents Policy generation, Opponent risk assessment and Environmental situational awareness.
With S-POE, three sub-benchmarks are accordingly designed, i.e., Multimodal Situational Awareness Benchmark (\textbf{MM-SA-Bench}), Psychological Reasoning – Opponent Modeling Benchmark (\textbf{PsyR-OM-Bench}), and Policy Generation for Gaming Benchmark (\textbf{PGG-Bench}).
The benchmark systematically assesses both state-of-the-art large models and human populations through carefully designed test samples.
Through comparative analysis, we identify current bottlenecks in large models' strategic reasoning and delineate critical pathways toward advancing strategic intelligence.

The rest of the paper is organized as follows.
In Section 2, we briefly describe benchmark's overall architecture and data provenance.
In sections 3-5, we elaborate on the three sub-benchmarks, i.e., MM-SA-Bench, PsyR-OM-Bench and PGG-Bench, and the evaluation results between large language models and human volunteers.
In Section 6, we introduce an LLM-based wargame agent for a comprehensive strategy reasoning assessment.
Finally, we conclude the paper in Section 7.

\section{Framework and Data description}
The overall framework of WGSR-Bench is displayed in Figure \ref{framework}.
MM-SA-Bench focuses on three tasks: object recognition, spatial relationship recognition, and situational reasoning analysis. It is further divided into 7 subtasks, including environmental identification, friend-foe positional relationships, and advantage/disadvantage assessment, comprising a total of 424 Q\&A pairs.
PsyR-OM-Bench revolves around two tasks: risk and reward. Designed with a three-tier structure—psychological traits, decision-making types, and behavioral manifestations—it includes 420 Q\&A pairs to evaluate adversary identification capabilities.
PGG-Bench centers on four typical game-theoretic tasks: non-cooperative games, incomplete information games, sequential games, and cooperative games. It is subdivided into 29 subtasks, encompassing deceptive tactical decision-making, anticipatory decision-making, and others, totaling 364 Q\&A pairs.

All the aforementioned Q\&A data is sampled from the MiaoSuan platform, which employs wargame simulations as its core framework. The platform brings together a large number of human participants and AI development teams to engage in both tournament-level and routine adversarial competitions.
The Q\&A pairs in WGSR-Bench are exclusively sourced from a real adversarial database, which contains over 400,000 battle replays.
The replays are approximately 2TB of textual data and more than 1PB of image/video data (generatable via replay rendering tools).
Key features of the data is:
\begin{itemize}
\item Data generation: The replays were generated through hybrid adversarial engagements involving tens of thousands of participants from over 7,000 organizations and more than 1,000 AI agents.
\item Scenario coverage: 225 scenarios across 6 major categories and 30 subcategories.
\item Scale Variability: Team sizes range from 3v3 to 30v30 entities, with dynamic entity counts allowing single-side engagements exceeding 100 entities.
\item Entity diversity: 5 major classes and 23 subclasses of entities.
\item Strategy length: Mission lengths vary from a few hundred steps up to 2,880 steps (maximum observed).
\end{itemize}

\begin{figure}
   \begin{center}
   \includegraphics[width=0.4875\textwidth]{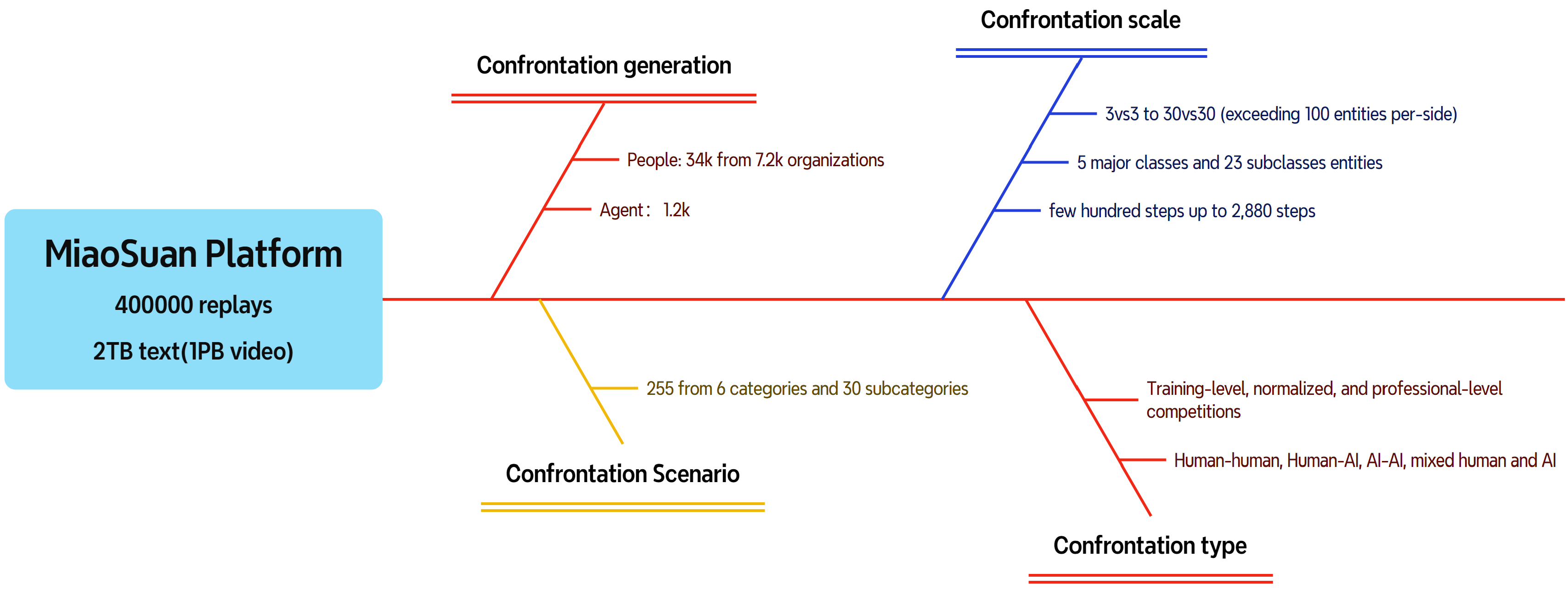}
   \end{center}
   \caption{Data used for WGSR-Bench.}
   \label{data}
\end{figure}

The large-scale real adversarial data ensures that WGSR-Bench covers a sufficiently broad scope, guaranteeing both high-quality test cases and the capacity for continuous expansion of the benchmark's scale.

To enable a comprehensive evaluation of large models, our selected models cover top-tier closed-source commercial models (e.g., GPT-4, Claude) and leading open-source models (e.g., Deepseek-R1, LLaMA).
As for human volunteers, We have selected about 150 personnel from professional colleges and universities across different expertise levels.
Regular-level participants consisted of entry-level strategists with basic game theory knowledge but limited practical experience. Superior-level participants were experienced players who demonstrated consistent strategic competence across multiple game types. Elite-level participants included national-level competitive players and military strategy instructors with deep domain expertise.

Finally, Large models are evaluated and compared with those volunteers for reasoning performance contrast.

\section{MM-SA-Bench}

\subsection{Overview}

In complex adversarial gaming environments, accurate situational understanding forms the prerequisite for strategy generation. To systematically evaluate this capability, we present the Multimodal Situational Awareness Benchmark (MM-SA-Bench), designed to assess large language models' ability to understand and reason about battlefield conditions. Situational Awareness (SA) serves not only as the starting point for decision-making but also as the foundational capability supporting high-level strategic reasoning. This is particularly crucial in wargame scenarios where participants must rapidly perceive battlefield conditions, identify key targets, analyze collaborative relationships, and assess situational trends under conditions of incomplete information and asymmetric actions.

Building upon established research in military cognition and battlefield modeling, situational awareness capabilities can be systematically decomposed into three core components. Target recognition involves perceiving critical terrain and force elements, forming the foundation of battlefield understanding. Relationship recognition requires understanding spatial and functional interdependencies between various battlefield entities. Comprehensive situation analysis synthesizes these elements into holistic battlefield assessment, including threat evaluation and control point dynamics. These three tasks form the cognitive loop that must be completed before any meaningful tactical decision-making can occur.

\subsection{Motivation and Challenges}

Current mainstream large language models have demonstrated strong capabilities in open-domain language generation, commonsense reasoning, and pattern learning. However, they exhibit significant limitations when confronted with the structured information perception, spatial reasoning, and multi-element collaborative understanding required in military contexts. Models struggle with systematic modeling of geometric relationships such as position, direction, and distance, often failing to correctly parse spatial descriptions like "Unit X is north of Point Y, forming an encirclement with Unit Z" that are commonplace in wargame scenarios.

The challenge extends beyond simple spatial understanding to encompass cross-modal information integration. While some multimodal models possess basic image-text understanding capabilities, achieving integrated analysis between battlefield visualizations and combat reports remains elusive. Furthermore, when facing complex collaborative relationships and multi-step scenarios typical of military operations, large models often fail to generate reasonable explanations or predict consequences, revealing fundamental gaps in their ability to process multi-causal chains. These limitations severely restrict the reliable application of current large language models in battlefield environments and create risks of "semantic detachment" where generated strategies lack grounding in tactical reality.

\subsection{Benchmark Design}

To systematically address these gaps and provide rigorous evaluation standards, we constructed MM-SA-Bench, a comprehensive benchmark focused on situational awareness tasks in wargame contexts. The benchmark draws from hundreds of thousands of real game replays from the MiaoSuan wargame platform, ensuring ecological validity and strategic complexity that reflects actual military decision-making scenarios.

MM-SA-Bench encompasses authentic complex data covering 6 major categories and 30 subcategories of typical wargame confrontation environments, providing rich diversity in tactical scenarios. The benchmark employs three task types derived from actual combat planning logic, systematically evaluating perception, comprehension, and strategic reasoning capabilities. All tasks utilize a unified multiple-choice format to ensure consistent evaluation and enable meaningful comparisons across different models and human participants. The design emphasizes cognitive boundary testing through information compression, ambiguity interference, and causal chain reasoning to challenge the limits of model understanding.

The benchmark evaluates performance across six core capability dimensions: Environment Recognition for assessing understanding of terrain and battlefield conditions, Operator Identification for testing ability to identify military units and assets, Inter-operator Relations for evaluating spatial reasoning between units, Operator-Keypoint Relations for assessing unit positions relative to strategic locations, Force Comparison for testing quantitative reasoning about relative strengths, and Complex Situation Analysis for evaluating high-level strategic assessment capabilities.

\subsection{Evaluation Results}

\subsubsection{Overall Performance Comparison}

\begin{figure}[h]
   \begin{center}
   \includegraphics[width=0.48\textwidth]{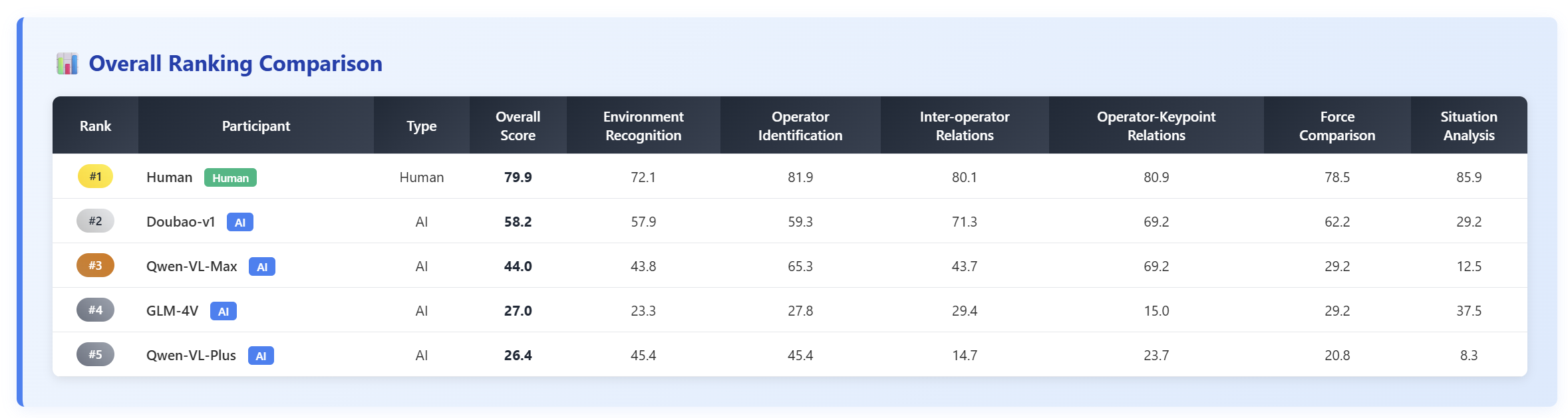}
   \end{center}
   \caption{Overall ranking comparison showing detailed performance breakdown across all participants in MM-SA-Bench}
   \label{fig:mm_sa_ranking}
\end{figure}

We conducted comprehensive evaluation with MM-SA-Bench, testing both human participants recruited from military academies and strategic gaming communities, alongside state-of-the-art multimodal language models. Figure~\ref{fig:mm_sa_ranking} presents the detailed performance breakdown across all participants.

The results reveal significant performance disparities between human experts and current AI systems. Human participants achieved an overall score of 79.9 points, demonstrating robust and consistent performance across diverse situational awareness tasks. In contrast, the best-performing AI model, Doubao-v1, achieved only 58.2 points, representing a substantial 21.7-point deficit. More concerning is the dramatic performance degradation observed in other models, with GLM-4V scoring 27.0 and Qwen-VL-Plus achieving only 26.4 points, less than one-third of human performance levels.

\begin{figure}[h]
   \begin{center}
   \includegraphics[width=0.45\textwidth]{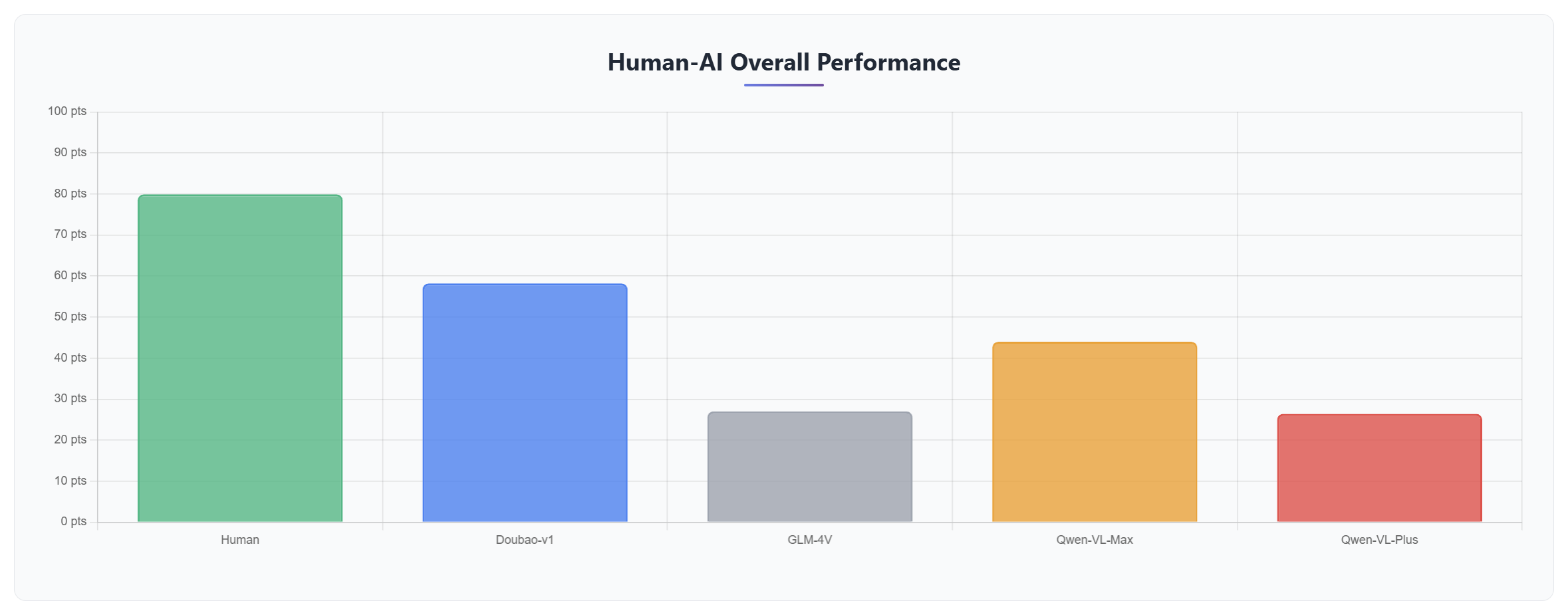}
   \end{center}
   \caption{Human-AI overall performance comparison in MM-SA-Bench}
   \label{fig:mm_sa_overall}
\end{figure}

Figure~\ref{fig:mm_sa_overall} provides a clear visualization of these performance disparities, illustrating the substantial gap between human and AI capabilities in situational awareness tasks. The stepwise performance degradation from Doubao-v1 to other models suggests that current architectural approaches face fundamental limitations in processing and understanding complex battlefield information.

\subsubsection{Capability-Specific Analysis}

\begin{figure}[h]
   \begin{center}
   \includegraphics[width=0.48\textwidth]{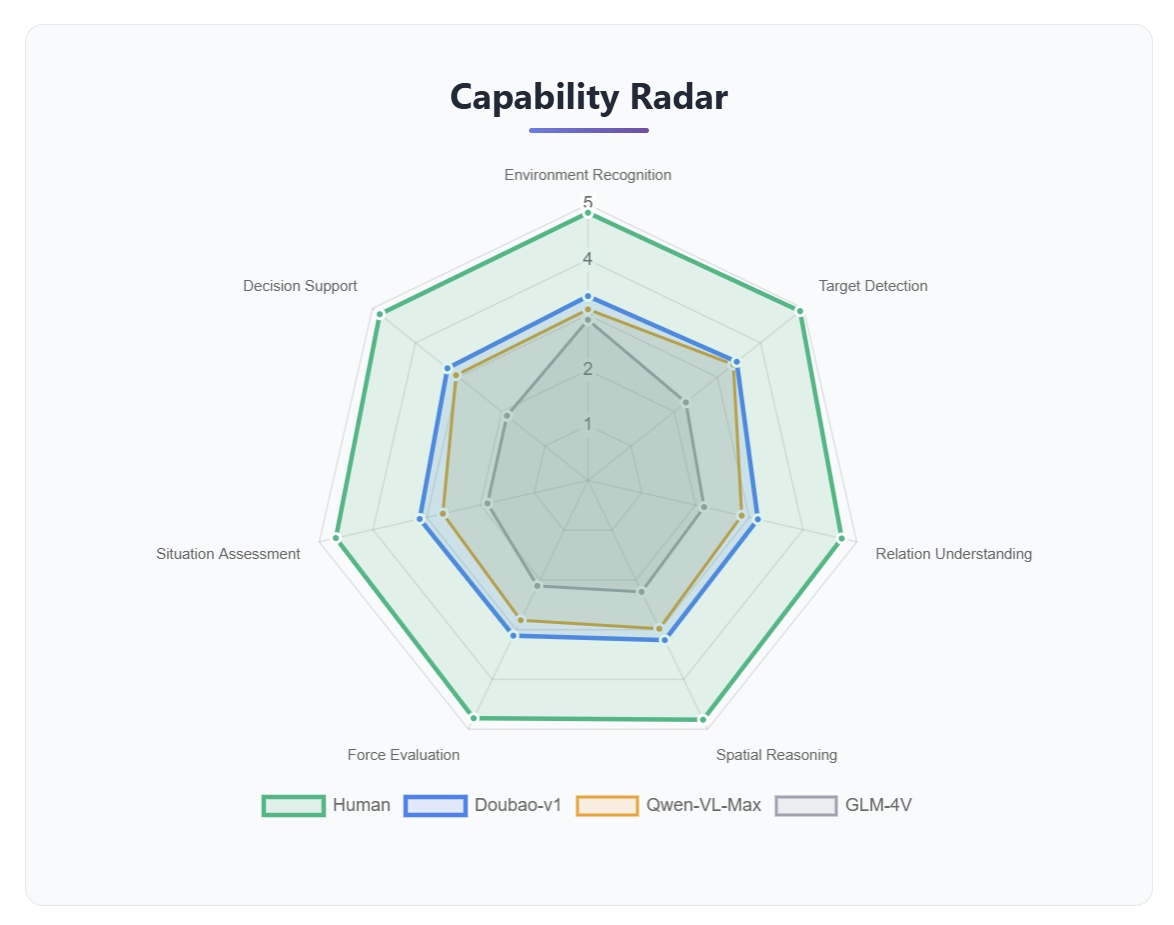}
   \end{center}
   \caption{Six core capability scores comparison across humans and AI models}
   \label{fig:mm_sa_capabilities}
\end{figure}

To understand the nature of these performance gaps more deeply, we analyzed model performance across the six core capability dimensions. Figure~\ref{fig:mm_sa_capabilities} presents a detailed breakdown of scores for each capability area, revealing nuanced patterns in model strengths and weaknesses.

The capability-specific analysis reveals that multimodal models demonstrate partially convergent performance with humans in certain spatial reasoning tasks. Notably, Doubao-v1 achieved 71.3\% accuracy in Inter-operator Relations compared to 80.1\% for humans, and 69.2\% in Operator-Keypoint Relations versus 80.9\% for humans. These results indicate that multimodal models have developed reasonably strong capabilities in specific spatial and directional judgment tasks, suggesting that basic geometric reasoning is within reach of current architectures.

However, the models exhibit severe deficits in high-level complex situation analysis tasks. While humans achieved 85.9\% accuracy in Complex Situation Analysis, the best-performing multimodal model (GLM-4V) managed only 37.5\%, revealing a striking 48.4 percentage point gap. This dramatic disparity indicates that multimodal models face fundamental limitations in high-dimensional abstract understanding and strategic synthesis, struggling to integrate multiple information sources into coherent battlefield assessments.

\begin{figure}[h]
   \begin{center}
   \includegraphics[width=0.45\textwidth]{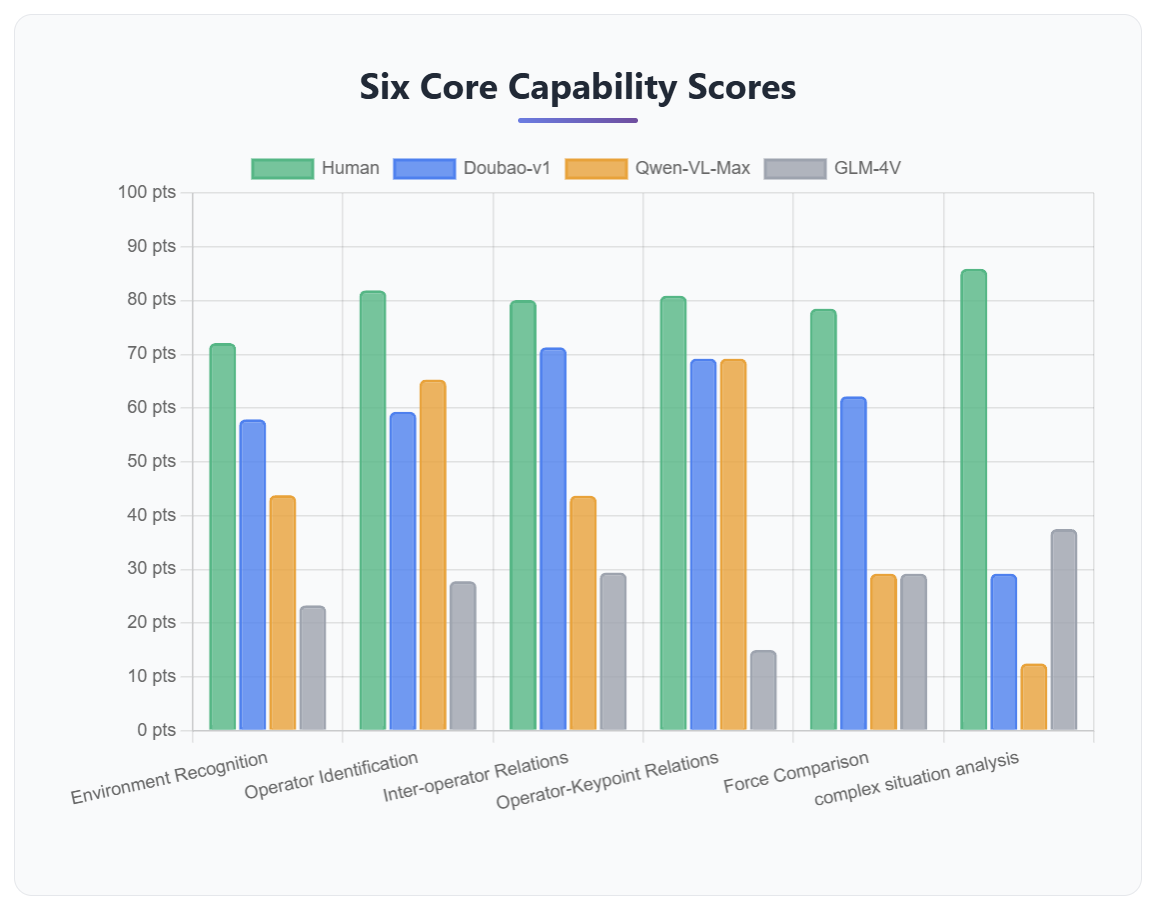}
   \end{center}
   \caption{Capability radar chart showing multi-dimensional performance profiles}
   \label{fig:mm_sa_radar}
\end{figure}

Figure~\ref{fig:mm_sa_radar} provides an alternative visualization through a radar chart, clearly illustrating the comprehensive superiority of human performance across all dimensions. The human performance profile forms a nearly perfect hexagon with minimal variation across capabilities, while AI models show severely distorted profiles with dramatic capability gaps and inconsistent performance patterns.

\subsection{Key Findings and Implications}

The evaluation results from MM-SA-Bench reveal significant differences between humans and multimodal large models in wargame situational awareness tasks. While multimodal models show partial capability convergence with humans in specific spatial reasoning tasks, particularly in inter-operator spatial relationships (80.1\% vs. 71.3\%) and operator-keypoint spatial relationships (80.9\% vs. 69.2\%), they face severe challenges in higher-order cognitive tasks. These results indicate that current models have developed reasonable capabilities for concrete spatial and directional judgments but struggle with abstract strategic reasoning.

The most pronounced performance gap emerges in complex situation analysis tasks, where humans achieved 85.9\% accuracy while the best multimodal model managed only 37.5\%. This substantial deficit demonstrates that multimodal models still face considerable challenges in high-dimensional abstract understanding compared to human capabilities. The gap is particularly concerning given that complex situation analysis represents the culmination of situational awareness, requiring synthesis of multiple information sources into actionable strategic insights.

Furthermore, multimodal models exhibit poor performance stability across different situational awareness tasks. While humans maintain consistent accuracy across all six task categories with an average of 79.9\% and variance of approximately ±10\%, multimodal models demonstrate pronounced "subject bias" with dramatic performance variations across different task types. This instability suggests that current models lack robust underlying representations for battlefield understanding and may be overfitting to specific task characteristics rather than developing generalizable situational awareness capabilities.

\subsection{Implications for Strategic AI Development}

The MM-SA-Bench evaluation results point to three concrete directions for improving strategic AI systems. First, the 48.4\% performance gap in complex situation analysis indicates that current models require specialized modules for multi-source information fusion and hierarchical reasoning. Second, the high variance in AI performance across different task types (ranging from 69.2\% to 12.5\% for Qwen-VL-Max) suggests that models need more balanced training data covering all aspects of battlefield understanding rather than optimizing for specific subtasks. Third, the relative success in spatial reasoning tasks (achieving over 70\% accuracy) demonstrates that targeted architectural improvements in these areas could yield immediate performance gains. These findings establish MM-SA-Bench as both a diagnostic tool and a development roadmap, providing quantitative targets for advancing AI capabilities in military strategic reasoning.

\section{PsyR-OM-Bench}

\subsection{Overview}

\par Unlike general reasoning tasks, game-theoretic strategy reasoning is inherently adversarial, meaning that the generation of our own strategy must explicitly account for the strategic characteristics of the opponent. Effective strategy formation under such settings is inseparable from the capacity to understand the opponent. This capacity is especially critical in highly uncertain and adversarial environments, such as those represented by war games, where the opponent’s optimal response is often non-unique. A deeper understanding of the opponent’s behavioral tendencies significantly enhances our ability to anticipate their strategic choices and proactively develop counter-strategies.

\par To rigorously evaluate the opponent modeling capabilities of large language models (LLMs), We introduce Psychological Reasoning – Opponent Modeling Benchmark (PsyR-OM-Bench), a novel evaluation framework designed to assess the ability of large language models to understand and model adversarial agents through latent psychological traits. This benchmark integrates key psychological traits known to influence decision-making—namely, risk preference and reward preference—and contextualizes them within a war-game setting to assess LLMs’ capacity for opponent understanding in high-uncertainty, high-conflict scenarios.

\par We decompose the opponent understanding task into two sub-tasks. The first sub-task is Risk-Reward Attribution, where the model is required to infer the underlying psychological preferences (i.e., risk-seeking vs. risk-averse; short-term vs. long-term reward orientation) of an agent based on partial behavioral trajectories. This requires the model not only to recognize the strategic implications of observed actions but also to map these behaviors to latent psychological traits that drive adversarial decisions. 

\par The second sub-task is Strategic Preference Prediction. In this task, LLMs are explicitly informed of the opponent’s risk and reward preferences. Based on this information, the models are required to select, from a set of multiple-choice options, the most plausible future strategic action the opponent is likely to take. This task evaluates the LLMs’ ability to integrate abstract psychological traits into concrete strategic reasoning. To succeed, models must go beyond surface-level language understanding and demonstrate several core capabilities.

\subsection{Motivation and Challenges}
\par Recent advances in large language models (LLMs) have demonstrated remarkable capabilities in abstract reasoning, theory of mind, and decision modeling across a wide range of natural language understanding tasks. LLMs have shown the ability to infer hidden intentions, generate plausible counterfactuals, and even simulate basic forms of social cognition. However, their effectiveness in adversarial strategic reasoning—particularly in modeling and anticipating the behavior of intelligent opponents in dynamic, high-stakes environments—remains largely underexplored. In real-world scenarios such as war games, financial negotiation, and cybersecurity defense, agents must contend not only with complex environments but also with strategic adversaries whose goals, beliefs, and reasoning processes may diverge significantly from their own. This necessitates a higher level of situational awareness, where the agent must go beyond observable actions to infer latent psychological traits that underpin adversarial decisions, such as risk tolerance, time preference, and reward orientation.

\par While LLMs excel in tasks grounded in factual knowledge and structured reasoning, adversarial domains introduce several unique and compounded challenges. First, the inference of psychological traits from behavioral traces often involves partial observability and ambiguity, where the same action may arise from distinct cognitive motivations. The model must therefore reason probabilistically and accommodate multiple explanatory hypotheses. Second, the translation from inferred traits to future action predictions requires dynamic simulation, contextual adaptation, and the integration of external constraints (e.g., game rules, environmental uncertainty, opponent modeling). Unlike static reasoning tasks, adversarial reasoning is iterative, interactive, and shaped by the anticipation of the other agent’s future moves, introducing higher-order belief reasoning and non-linear causal dependencies.

\par These challenges give rise to a natural asymmetry between the core subtasks of opponent modeling. The bottom-up process of trait inference from behavior may benefit more directly from language-based pattern matching and pretrained statistical associations. In contrast, the top-down process of action prediction from traits demands generative simulation under uncertainty, involving forward reasoning, goal inference, and counterfactual analysis. The cognitive demands of the latter more closely resemble human strategic planning, where decision-makers must construct mental models of their opponents and mentally project possible trajectories. Thus, a comprehensive evaluation of LLMs in this space must consider both directions of reasoning and the degree to which they can generalize across varying contexts and agent profiles.

\par Despite the centrality of these capabilities to high-level AI cognition, current benchmarks lack targeted tasks that assess trait-based strategic reasoning in adversarial settings. Existing opponent modeling datasets often assume full observability, fixed strategies, or disregard the role of psychological dispositions in decision-making. Consequently, there is a pressing need for benchmarks that bridge psychological theory and strategic game dynamics, enabling a deeper understanding of how LLMs represent, reason about, and simulate adversarial agents under uncertainty.

\par To fill this gap, we propose PsyR-OM-Bench—a psychologically grounded, adversarial reasoning benchmark that frames opponent modeling as a two-stage cognitive task: first, inferring latent traits from past actions; second, predicting future strategic moves from known traits. By decoupling and systematically evaluating these two components, PsyR-OM-Bench offers a structured framework for analyzing the strengths and limitations of LLMs in modeling adversarial behavior. Importantly, the benchmark emphasizes trait-action generalization, context-sensitive reasoning, and scenario simulation, all of which are essential capabilities for AI systems deployed in real-world competitive or collaborative multi-agent environments.

\subsection{Evaluation Results}

To assess the capabilities of large language models in psychologically grounded opponent modeling, we conducted a comprehensive evaluation using PsyR-OM-Bench across six representative agents: three human groups—elite-level, professional-level, and general-level participants—and three state-of-the-art LLMs—DeepSeek R1, Qwen3, and DeepSeek V3. The benchmark consists of two complementary sub-tasks: (1) Risk-Reward Attribution, which requires inferring psychological traits from behavioral sequences, and (2) Strategic Preference Prediction, which evaluates the model’s ability to predict future decisions based on given trait profiles. Each sub-task is further broken down into four key psychological categories: high risk, low risk, long-term reward, and short-term reward, enabling fine-grained performance assessment.

\begin{figure}[!tbp]
   \begin{center}
   \includegraphics[width=0.48\textwidth]{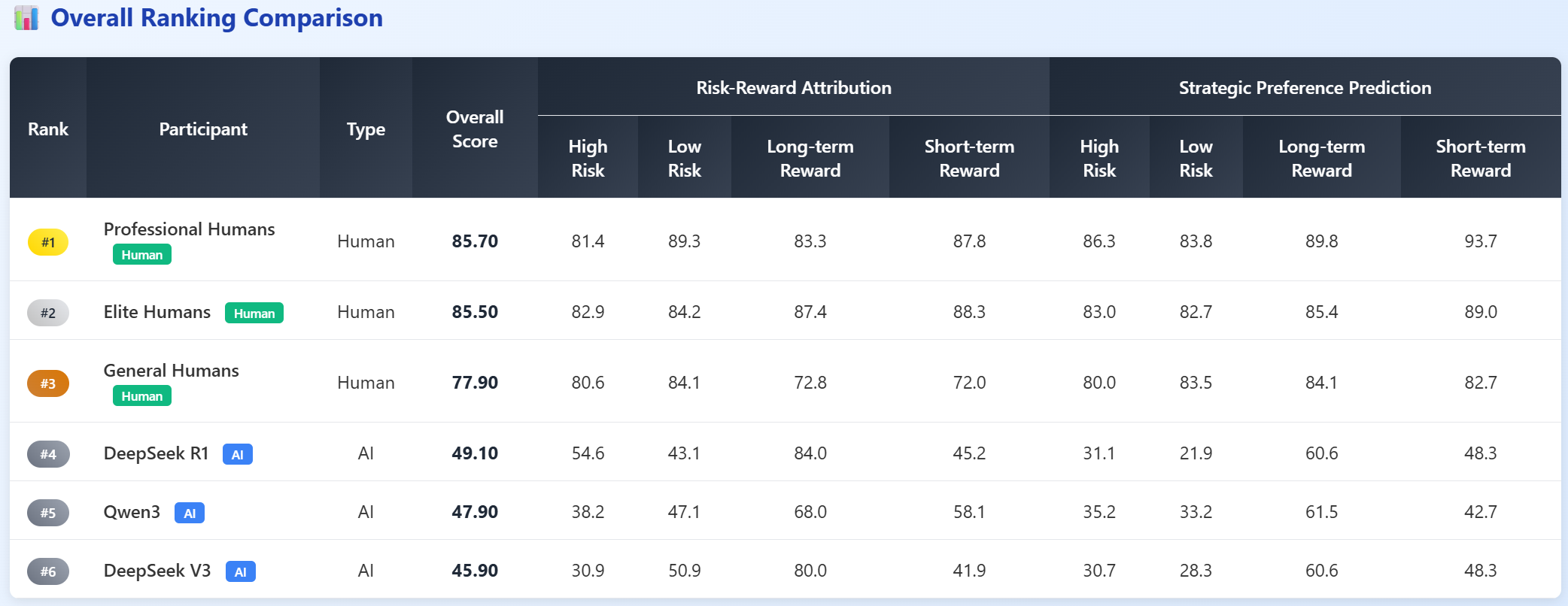}
   \end{center}
   \caption{Overall ranking comparison showing detailed performance breakdown across all participants in PsyR-OM-Bench.}
   \label{fig:PsyR_ranking}
\end{figure}

\subsubsection{Overall Performance Comparison}
\par
\begin{figure}[!tbp]
   \begin{center}
   \includegraphics[width=0.45\textwidth]{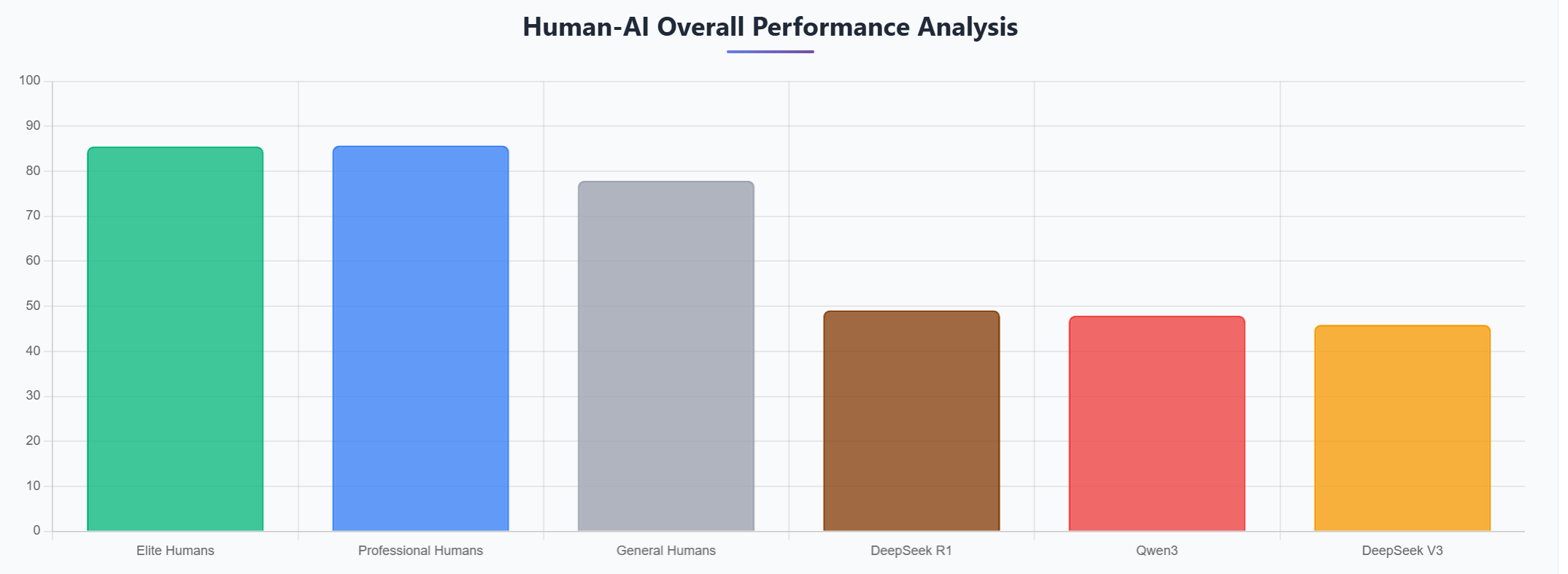}
   \end{center}
   \caption{Human-AI overall performance comparison in PsyR-OM-Bench}
   \label{fig:pysr_om_overall}
\end{figure}

 The overall performance ranking across all six evaluated entities—three human groups and three large language models—reveals a clear and consistent hierarchy of strategic reasoning ability. Figure~\ref{fig:PsyR_ranking} and Figure~\ref{fig:pysr_om_overall} present the detailed performance breakdown across all participants. Human participants occupy the top three positions, with a pronounced gap separating them from even the best-performing LLMs. The professional-level human group achieved the highest composite score (85.7), closely followed by the elite-level group (85.5). These two groups displayed near-ceiling performance across most dimensions, indicating not only a deep understanding of opponent behavior but also a high degree of cognitive flexibility across diverse psychological profiles and task complexities. 

\par The general-level human group, despite being composed of individuals without professional training in game theory or psychology, attained an impressive score of 77.9, significantly outperforming all tested LLMs. This result underscores the strength of intuitive human reasoning and social cognition in adversarial contexts, even among non-expert populations.

\par Among the LLMs, DeepSeek R1 ranked highest with a total score of 49.1, followed by Qwen3 at 47.9 and DeepSeek V3 at 45.9. While these scores are above random baseline and reflect some capacity for structured reasoning, they fall well short of human performance, particularly in tasks requiring forward planning or nuanced psychological interpretation.

\par Interestingly, the score differential between the top LLM (DeepSeek R1) and the lowest-ranking human group (general-level) is nearly 29 points, indicating that current LLMs remain at least one generation behind in their ability to model strategic cognition at a human-comparable level. This finding persists despite the fact that these LLMs are among the most capable open-access models, incorporating billions of parameters and trained on diverse datasets.

\par A closer inspection of the composite scores also reveals differential consistency. Human groups not only performed better on average, but also exhibited lower variance across dimensions and sub-tasks. In contrast, LLMs demonstrated irregular performance profiles: certain psychological categories—such as long-term reward attribution—were handled with moderate success, while others—especially high-risk strategy prediction—saw dramatic performance drops. This suggests that LLMs may lack a unified internal representation for psychological state modeling and instead rely on fragmented heuristics.

\par The ranking structure also implies potential areas of confusion or misrepresentation within the models. For example, Qwen3, while generally weaker in attribution tasks compared to DeepSeek R1, slightly outperformed it in long-term reward prediction, indicating the presence of localized reasoning strengths possibly tied to its training corpus or decoding behavior. However, such instances were not sufficient to offset weaknesses in high-variance or ambiguous scenarios.

\par Moreover, the sharp divide between human and AI scores was preserved even when aggregating across risk and reward categories, suggesting that the observed gaps are not isolated anomalies but instead systemic limitations of current LLM architectures when applied to opponent modeling tasks requiring abstract, theory-of-mind-based reasoning.

\par To summarize, the overall comparison underscores a robust and consistent performance advantage for human participants—especially trained professionals—across all evaluation dimensions. The current generation of LLMs, despite excelling in general-purpose NLP tasks, remains substantially constrained in their ability to internalize, simulate, and anticipate human-like strategic behavior in adversarial environments. This gap motivates the need for more targeted training regimes and evaluation protocols aimed at closing the human-AI reasoning divide in psychologically grounded, high-stakes decision-making tasks.

\subsubsection{Sub-task Performance Breakdownn}
\par
\begin{figure}[!tbp]
   \begin{center}
   \includegraphics[width=0.48\textwidth]{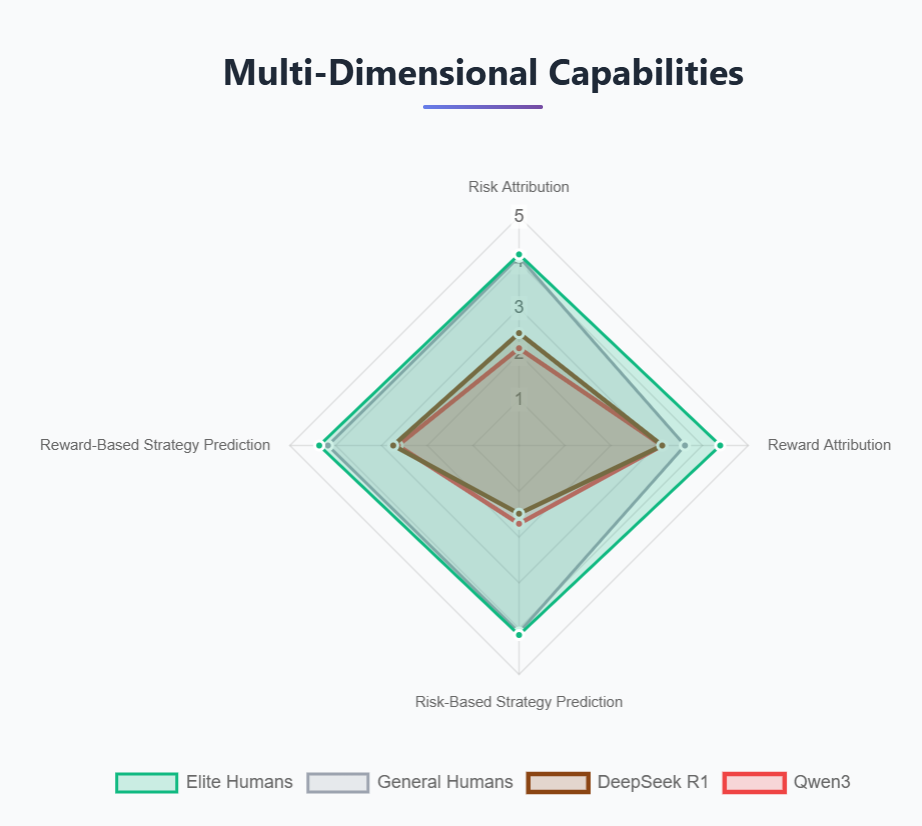}
   \end{center}
   \caption{Four core capability scores comparison across humans and AI models}
   \label{fig:pysr_om_capabilities}
\end{figure}

\begin{figure}[!tbp]
   \begin{center}
   \includegraphics[width=0.45\textwidth]{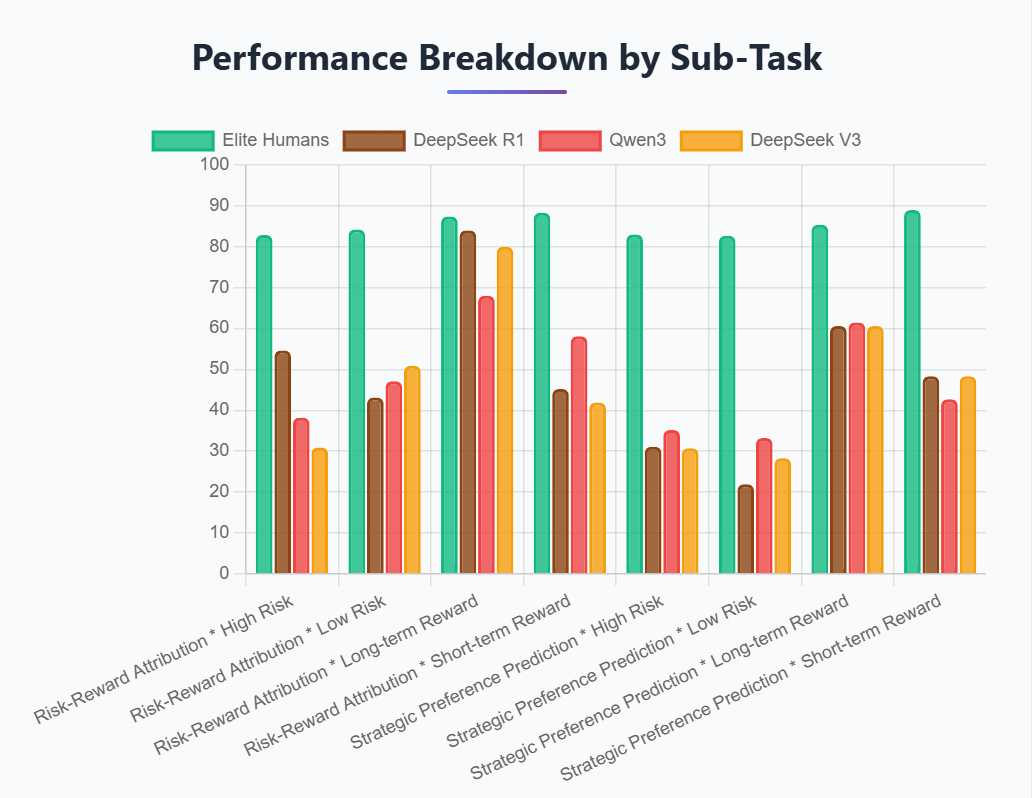}
   \end{center}
   \caption{Performance across different subtasks showing task-specific strategic reasoning capabilities}
   \label{fig:pysr_om_subtask}
\end{figure}

 To gain deeper insight into the reasoning capabilities of humans and LLMs under PsyR-OM-Bench, we further decompose the evaluation into its two core sub-tasks: Risk-Reward Attribution and Strategic Preference Prediction. These sub-tasks probe complementary dimensions of cognitive reasoning: the former emphasizes bottom-up inference from observed behavior to internal traits, while the latter requires top-down prediction from psychological priors to likely actions under uncertainty. 
Figure~\ref{fig:pysr_om_capabilities} and Figure~\ref{fig:pysr_om_subtask} presents a detailed breakdown of scores for each subtask.

\textbf{Risk-Reward Attribution.}  In the attribution task, human participants demonstrated high and stable performance across all psychological categories. The elite-level and professional-level groups consistently achieved scores above 80 in all four dimensions—high risk, low risk, long-term reward, and short-term reward—with professional participants reaching a peak of 89.3 in low-risk attribution and 87.8 in short-term reward attribution. These results suggest that trained humans possess not only domain knowledge but also a finely calibrated ability to interpret behavioral cues and map them to latent cognitive profiles. General-level humans, while slightly lower, still maintained solid performance (72.0–84.1) across categories, particularly excelling in risk sensitivity. This illustrates the robustness of intuitive social reasoning and trait attribution even in the absence of formal training. In contrast, LLMs exhibited highly uneven attribution performance. The most striking feature was their stronger ability in long-term reward attribution: for example, DeepSeek R1 scored 84.0, and DeepSeek V3 scored 80.0, approaching human-level performance in this specific subdimension.

\par However, LLMs faltered in short-term reward attribution, with DeepSeek R1 and DeepSeek V3 scoring only 45.2 and 41.9, respectively. This indicates difficulty in interpreting opportunistic or reactive behavior, which often lacks long-range coherence and requires fine-grained contextual understanding. Similarly, attribution in high-risk scenarios posed significant challenges: Qwen3 scored 38.2, and DeepSeek V3 scored 30.9, suggesting LLMs have difficulty distinguishing risk-seeking behavior from mere randomness, particularly when action sequences are short or ambiguous.

\par Interestingly, some models (e.g., Qwen3) performed slightly better in low-risk attribution (47.1) than in high-risk, indicating possible calibration around normative behavior but not deviations from it. This asymmetry may be rooted in the fact that most LLM training corpora are dominated by cautious, normative behavioral patterns, leading to poor generalization when faced with aggressive or unorthodox strategies.

\textbf{Strategic Preference Prediction.} The prediction sub-task is cognitively more demanding, requiring the model to reason in a top-down manner: given a profile of psychological traits, the model must simulate a plausible decision that an adversarial agent would make in a novel context.

Professional-level human participants excelled in this task, with all sub-dimensions scoring above 83, and achieving 93.7 in short-term reward prediction—the highest single-dimension score across the entire benchmark. This suggests a well-developed capacity for simulating other agents' behavior based on abstract traits, as well as flexible adaptation to different scenario types. Elite-level humans also performed at a high level (82–89 range), with slightly lower but still consistent results. General-level humans remained strong performers (80.0–84.1), indicating that basic trait-to-action simulation is accessible even without domain-specific expertise. Their balanced scores across risk and reward conditions also suggest that this form of reasoning is cognitively intuitive for many individuals, perhaps reflecting natural theory-of-mind abilities.

LLMs, by contrast, showed marked performance degradation in this task. Their risk-based predictions were the weakest: for example, DeepSeek R1 scored 31.1 in high-risk strategy prediction, and Qwen3 scored 35.2, suggesting that LLMs struggled to simulate risk-seeking behavior even when the opponent’s profile was explicitly provided. This reflects an inability to integrate psychological priors into dynamic decision-making in volatile settings. LLMs fared slightly better in long-term reward prediction (e.g., DeepSeek R1 and V3 both scored 60.6), again confirming their relative comfort in temporally consistent planning domains. However, their performance in short-term reward strategy prediction remained low (48.3 for R1; 42.7 for Qwen3), further reinforcing the conclusion that LLMs are under-equipped to reason about fast, local optimizations or impulsive actions.

Notably, the performance variance across LLMs in this sub-task was greater than in attribution, suggesting that small differences in architecture or training corpus can lead to disproportionately large effects in top-down reasoning tasks. For example, despite being trained on similar data, Qwen3 and DeepSeek R1 differed by over 12 points in some sub-dimensions, highlighting the sensitivity of strategic simulation to model configuration.

\textbf{Cross-Task Observations. } The divergence in LLM performance across the two sub-tasks suggests that their internal representations may not yet support bi-directional cognitive integration. Human participants appear to leverage inferred traits to consistently inform their predictions, while LLMs seem to treat attribution and prediction as loosely coupled tasks, lacking the ability to transfer inferred psychological knowledge to downstream planning decisions.

This breakdown is particularly evident in risk-based dimensions, where LLMs often inferred traits at near-random levels and subsequently failed to use them in strategic forecasting. Conversely, their stronger performance in long-term reward prediction may reflect over-reliance on default planning priors learned from language modeling (e.g., preferring patient, deterministic strategies).

\subsection{Key Findings and Implications}

\textbf{Human Cognition Significantly Outperforms LLMs Across All Dimensions.}  All three human groups—elite, professional, and general-level—consistently outperformed large language models across both attribution and prediction sub-tasks. Even general-level humans outscored all LLMs by a wide margin, reaffirming that current LLMs lack the depth of psychological reasoning, strategic flexibility, and adversarial adaptability required for high-fidelity opponent modeling.

\textbf{LLMs Exhibit a Reasoning Asymmetry: Stronger in Induction, Weaker in Deduction.}  The evaluation reveals a structural imbalance in LLM reasoning: models such as DeepSeek R1 performed moderately well in inductive tasks like trait attribution (e.g., risk recognition score: 55.8), but fell short in deductive reasoning tasks like strategy prediction (42.2, vs. human: 88.5). This pattern suggests that LLMs excel at pattern recognition but struggle with causal inference and dynamic simulation, which are essential for forecasting strategic behaviors under uncertainty.

\textbf{Limited Generalization Under Semantic Ambiguity.}  While LLMs showed relatively stable performance in reward-based tasks with clear semantic framing (score: 59.7), their accuracy dropped sharply in risk-based tasks involving ambiguous or vague language (e.g., “moderately high risk”), with an average score of 39.5. In contrast, human participants handled both categories evenly (88.7 vs. 85.2), suggesting that humans better navigate linguistic imprecision and contextual nuance, while LLMs are overly reliant on surface-level semantic clarity.

\textbf{Temporal Reasoning is Biased Toward Long Horizons.}  LLMs demonstrated significantly better performance in long-term reward modeling (e.g., DeepSeek R1: 84.0, close to human: 83.3), but failed in short-term scenarios (DeepSeek R1: 45.2 vs. human: 87.8). This discrepancy likely stems from two factors: (i) pretraining biases that favor temporally extended planning, and (ii) lack of support for combinatorial reasoning under short-horizon conditions, where decision spaces are larger and trade-offs more acute.

\textbf{LLMs Struggle with High-Risk Strategic Scenarios.} Across both sub-tasks, LLMs consistently underperformed in high-risk conditions—for instance, scoring below 40 in both risk attribution and risk-based strategy prediction. These scenarios demand volatility modeling, probabilistic inference, and adversarial intent recognition—abilities that remain largely out of reach for current LLMs. Human participants, conversely, maintained strong and balanced performance in both high- and low-risk settings.

\textbf{PsyR-OM-Bench Reveals Critical Gaps in LLM Cognition.} Unlike generic NLP benchmarks, PsyR-OM-Bench targets trait-to-strategy reasoning, exposing specific cognitive weaknesses in LLMs. The benchmark’s decomposition into interpretable psychological dimensions—such as risk preference, reward horizon, and decision ambiguity—enables fine-grained diagnostic evaluation. This underscores the necessity of trait-driven evaluation frameworks that go beyond surface-level correctness to assess deeper reasoning fidelity.

\textbf{Toward Human-AI Collaborative Strategy Systems.} Despite their limitations, LLMs showed localized strengths—particularly in trait inference under clear conditions and long-term preference modeling. These partial competencies suggest a hybrid design direction: LLMs can serve as trait summarization and pattern recognition assistants, while humans handle adaptive planning, ambiguity resolution, and high-risk simulation. Future human-in-the-loop architectures may effectively combine symbolic reasoning, language modeling, and experiential decision support.

\section{PGG-Bench}

\subsection{Overview}

The Policy Generation for Gaming Benchmark (PGG-Bench) constitutes the third pillar of WGSR-Bench, specifically designed to evaluate strategic policy generation capabilities of large language models in complex wargaming scenarios. This benchmark addresses a critical gap in existing evaluation frameworks by systematically assessing models' abilities to formulate, adapt, and execute strategic policies under adversarial conditions with incomplete information.

\subsection{Benchmark Architecture}

PGG-Bench, representing the Policy generation component within the broader S-POE framework of WGSR-Bench, is grounded in combinatorial game theory and encompasses four fundamental game-theoretic paradigms: non-cooperative games, incomplete information games, sequential games, and cooperative games. The benchmark systematically evaluates strategic policy generation capabilities through 28 distinct decision types instantiated across 364 carefully designed strategic question-answer pairs. This comprehensive coverage ensures evaluation of models' abilities to handle diverse strategic contexts, from zero-sum competitive scenarios to complex coalition formation under uncertainty.

As the policy generation component of WGSR-Bench, PGG-Bench specifically focuses on evaluating how models synthesize situational awareness and opponent modeling into actionable strategic policies. This focus on policy generation represents the culmination of strategic reasoning, where understanding of the environment and adversary must be transformed into concrete action sequences. The benchmark tests this synthesis capability across varying temporal horizons and complexity levels, from immediate tactical responses to long-term campaign strategies spanning hundreds of moves.

\subsection{Evaluation Methodology}

\subsubsection{Task Design}

PGG-Bench utilizes a multimodal approach that seamlessly integrates visual game states with textual strategic queries. Each evaluation instance presents participants with a hexagonal grid-based wargame visualization displaying unit positions, terrain features, and strategic assets. This visual information is complemented by rich contextual details including force compositions, recent move histories, and environmental constraints that shape the strategic landscape. Participants must then respond to multiple-choice questions requiring strategic analysis and decision-making, as well as free-form policy generation tasks demanding comprehensive strategic planning.

The benchmark encompasses 28 distinct decision types distributed across 364 strategic scenarios, ensuring comprehensive coverage of strategic reasoning dimensions. These tasks span a wide spectrum from immediate tactical decisions requiring rapid situational assessment to long-horizon strategic planning demanding sustained coherent reasoning. The complexity levels are carefully calibrated to differentiate between novice and expert-level strategic reasoning, with each scenario incorporating varying degrees of uncertainty, adversarial dynamics, and strategic depth.

\subsubsection{Automatic Evaluation Framework}

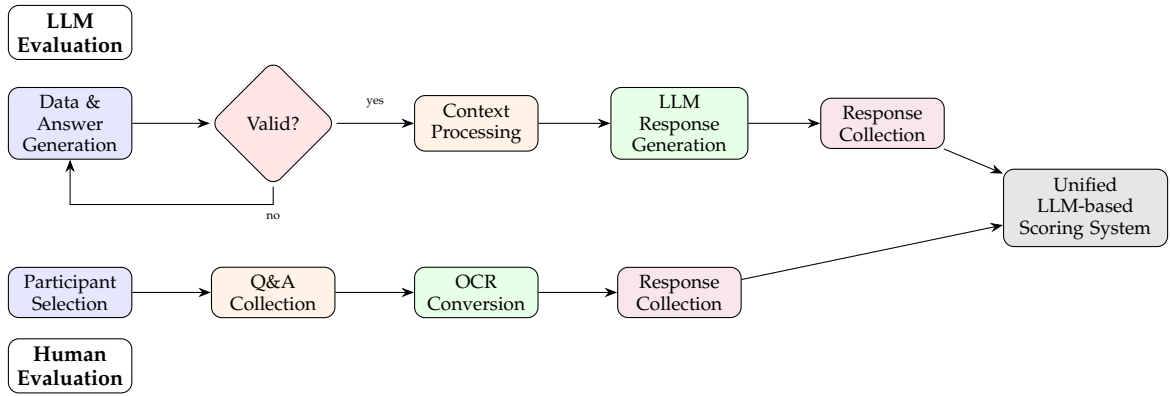
\begin{figure*}[t]
\centering
\begin{tikzpicture}[
    every node/.style={font=\footnotesize,draw,rounded corners,minimum width=1.6cm,minimum height=0.6cm,text width=1.6cm,align=center},
    node distance=2.2cm and 2.8cm,
    >=Stealth,
    scale=0.9,
    transform shape
]

\node[fill=blue!10] (llm_start) at (0,2.5) {Data \& Answer Generation};
\node[diamond,fill=red!10,minimum width=1.2cm,minimum height=1cm,text width=1.2cm] (llm_check) at (3,2.5) {Valid?};
\node[fill=orange!10] (llm_process) at (6,2.5) {Context Processing};
\node[fill=green!10,text width=1.8cm] (llm_gen) at (9,2.5) {LLM Response Generation};
\node[fill=purple!10] (llm_collect) at (12,2.5) {Response Collection};

\node[fill=blue!10] (human_select) at (0,0) {Participant Selection};
\node[fill=orange!10] (human_qa) at (3,0) {Q\&A Collection};
\node[fill=green!10] (human_ocr) at (6,0) {OCR Conversion};
\node[fill=purple!10] (human_format) at (9,0) {Response Collection};

\node[fill=gray!20,minimum width=2.2cm,minimum height=0.8cm,text width=2.2cm] (unified_scorer) at (15,1.25) {Unified LLM-based Scoring System};

\draw[->] (llm_start) -- (llm_check);
\draw[->] (llm_check) -- node[above,font=\tiny,draw=none] {yes} (llm_process);
\draw[->] (llm_check) -- node[below,font=\tiny,draw=none] {no} +(0,-1.2) -- +(-3,-1.2) -- (llm_start);
\draw[->] (llm_process) -- (llm_gen);
\draw[->] (llm_gen) -- (llm_collect);

\draw[->] (human_select) -- (human_qa);
\draw[->] (human_qa) -- (human_ocr);
\draw[->] (human_ocr) -- (human_format);

\draw[->] (llm_collect) -- (unified_scorer);
\draw[->] (human_format) -- (unified_scorer);

\node[above=0.4cm of llm_start,font=\small] {\textbf{LLM Evaluation}};
\node[below=0.3cm of human_select,font=\small] {\textbf{Human Evaluation}};
\end{tikzpicture}
\caption{Unified evaluation process for PGG-Bench: identical scenarios presented to both humans and AI, with responses fed into a single LLM-based scoring system}
\label{fig:eval_process}
\end{figure*}

To ensure scalable and consistent evaluation, PGG-Bench employs an automatic scoring framework powered by advanced language models. The evaluation process, illustrated in Figure~\ref{fig:eval_process}, presents identical strategic scenarios to both human participants and AI models, collecting their responses separately, then feeding both response sets into a unified LLM-based evaluation system for consistent scoring and comparison.

\subsubsection{Theoretical Foundation for Assessment Standards}

Our evaluation framework is grounded in robust educational assessment theory, integrating principles from both educational evaluation and evaluation research domains. Drawing from Anderson and Krathwohl's work \cite{anderson2001revision} on assessment design, we incorporate structured standards and descriptors to ensure effective evaluation of complex, open-ended tasks. The framework leverages authentic assessment principles \cite{gulikers2004five}, which emphasize real-world relevance and performance-based evaluation, particularly suitable for strategic reasoning tasks that mirror actual decision-making scenarios.

Our approach combines educational evaluation theory with consensus principles from the assessment field \cite{brookhart2018appropriate}. We employ evaluation rubrics—systematic scoring guides that articulate specific performance criteria—following Popham's framework\cite{popham2000modern} for constructed response evaluation. This methodology ensures assessment of multidimensional capabilities across different performance levels, addressing the inherent complexity of strategic policy generation \cite{reddy2010review}.

The assessment framework specifically incorporates authentic assessment methodologies, which are particularly relevant for evaluating strategic reasoning
 as they assess students' ability to apply knowledge in realistic contexts. This aligns with Wiggins's emphasis \cite{wiggins1998educative} on performance tasks that require complex thinking and application of knowledge, rather than mere recall or recognition.

\subsubsection{Methodological Alignment with Bloom's Taxonomy}

To ensure methodological rigor, our evaluation dimensions are systematically aligned with the revised Bloom's Taxonomy of Educational Objectives \cite{anderson2001revision}, providing a solid educational foundation for our assessment framework. This alignment ensures comprehensive evaluation across cognitive complexity levels and supports the development of higher-order thinking skills assessment \cite{krathwohl2002revision}.

Our evaluation dimensions correspond to specific cognitive levels:

\begin{itemize}
\item \textbf{Factual Correctness} aligns with the \textit{Remember} level, assessing accurate recall and recognition of game state information, rules, and strategic principles.
\item \textbf{Logical Consistency} corresponds to the \textit{Understand} level, evaluating the ability to construct meaning from strategic information and demonstrate comprehension of cause-effect relationships.
\item \textbf{Game Principles Adherence} relates to the \textit{Apply} and \textit{Analyze} levels, measuring the ability to use strategic knowledge in new situations and examine relationships between strategic elements.
\item \textbf{Outcome Prediction} encompasses the \textit{Evaluate} and \textit{Create} levels, assessing the capability to make judgments about strategic alternatives and synthesize information into coherent strategic plans.
\item \textbf{Clarity and Completeness} reflects \textit{Create} level abilities, evaluating the synthesis and articulation of comprehensive strategic reasoning.
\end{itemize}

This taxonomic alignment ensures that our evaluation captures the full spectrum of cognitive processes required for strategic reasoning, from basic knowledge recall to complex strategic synthesis \cite{bloom1956taxonomy}. The framework supports Gosselin and Okamoto's recommendations\cite{gosselin2018improving} for using Bloom's taxonomy to improve assessment alignment and ensure evaluation of higher-order learning.

\subsubsection{Automated Scoring Implementation}

The automatic scoring system employs a sophisticated six-dimensional evaluation framework that extends beyond traditional five-dimensional approaches by incorporating innovation assessment. The evaluation leverages Qwen-2.5-VL-72B as the primary scoring model, following recent advances in LLM-based assessment \cite{hashemi2024llm}. The scoring process utilizes structured JSON output formats to enable systematic decomposition and analysis of strategic responses, facilitating both automated processing and detailed qualitative analysis.

The JSON-structured output serves a critical role in our evaluation methodology, enabling what Fu et al. \cite{fu2023gptscore} term ``structured evaluation decomposition.'' This approach allows for granular analysis of different aspects of strategic reasoning while maintaining consistency across evaluations. The structured format supports both automated analysis and human verification, addressing the reliability concerns raised by Liu et al. \cite{liu2023evaluating} regarding LLM-based evaluation systems.

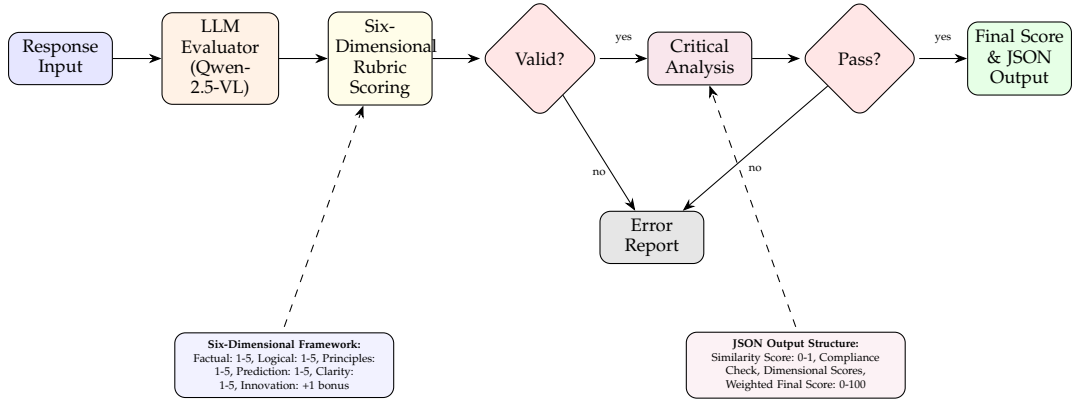
\begin{figure*}[t]
\centering
\begin{tikzpicture}[
    every node/.style={font=\footnotesize,draw,rounded corners,minimum width=1.4cm,minimum height=0.6cm,text width=1.4cm,align=center},
    node distance=1.8cm and 2.2cm,
    >=Stealth,
    scale=0.85,
    transform shape
]

\node[fill=blue!10] (input) at (0,2) {Response Input};
\node[fill=orange!10,text width=1.6cm] (llm_eval) at (2.5,2) {LLM Evaluator (Qwen-2.5-VL)};
\node[fill=yellow!10] (rubric) at (5,2) {Six-Dimensional Rubric Scoring};
\node[diamond,fill=red!10,minimum width=1.2cm,minimum height=1cm,text width=1.2cm] (check) at (7.5,2) {Valid?};
\node[fill=purple!10] (critical) at (10,2) {Critical Analysis};
\node[diamond,fill=red!10,minimum width=1.2cm,minimum height=1cm,text width=1.2cm] (final_check) at (12.5,2) {Pass?};
\node[fill=green!10] (output) at (15,2) {Final Score \& JSON Output};

\node[fill=gray!20] (error) at (9.25,-0.8) {Error Report};

\draw[->] (input) -- (llm_eval);
\draw[->] (llm_eval) -- (rubric);
\draw[->] (rubric) -- (check);
\draw[->] (check) -- node[above,font=\tiny,draw=none] {yes} (critical);
\draw[->] (check) -- node[below,font=\tiny,draw=none] {no} (error);
\draw[->] (critical) -- (final_check);
\draw[->] (final_check) -- node[above,font=\tiny,draw=none] {yes} (output);
\draw[->] (final_check) -- node[below,font=\tiny,draw=none] {no} (error);

\node[draw, rounded corners, fill=blue!5, text width=3.2cm, align=center,font=\tiny] at (3.5,-2.8) {
\textbf{Six-Dimensional Framework:} Factual: 1-5, Logical: 1-5, Principles: 1-5, Prediction: 1-5, Clarity: 1-5, Innovation: +1 bonus
};

\node[draw, rounded corners, fill=purple!5, text width=3.2cm, align=center,font=\tiny] at (11.5,-2.8) {
\textbf{JSON Output Structure:} Similarity Score: 0-1, Compliance Check, Dimensional Scores, Weighted Final Score: 0-100
};

\draw[dashed, ->] (3.5,-2.2) -- (rubric);
\draw[dashed, ->] (11.5,-2.2) -- (critical);
\end{tikzpicture}
\caption{Six-dimensional automatic scoring pipeline with structured JSON evaluation output}
\label{fig:scoring_flow}
\end{figure*}

Figure~\ref{fig:scoring_flow} illustrates the detailed scoring pipeline, where responses undergo multi-stage evaluation culminating in a comprehensive score.

\textbf{Six-Dimensional Evaluation Framework}

Our evaluation system extends traditional rubric-based assessment by implementing a comprehensive six-dimensional framework:

\begin{itemize}
\item \textbf{Factual Correctness} (1-5 scale): Accuracy of game state interpretation, rule application, and factual claims about strategic situations.
\item \textbf{Logical Consistency} (1-5 scale): Coherence of reasoning chains, internal consistency of arguments, and logical validity of strategic justifications.  
\item \textbf{Game Principles Adherence} (1-5 scale): Compliance with established strategic principles, game-theoretic concepts, and domain-specific best practices.
\item \textbf{Outcome Prediction} (1-5 scale): Quality of consequence analysis, accuracy of future state projections, and understanding of strategic implications.
\item \textbf{Clarity and Completeness} (1-5 scale): Comprehensiveness of strategic reasoning, articulation quality, and completeness of strategic plans.
\item \textbf{Innovation Bonus} (+1 point): Recognition of creative or novel strategic approaches that demonstrate advanced strategic thinking while maintaining validity.
\end{itemize}

This framework addresses limitations identified in previous reviews of automated assessment systems by providing explicit criteria for evaluating creative and innovative responses. The innovation dimension acknowledges that effective strategic reasoning often requires novel approaches, addressing Wang et al. \cite{wang2023evaluation} call for assessment frameworks that recognize valid alternative solutions in complex reasoning tasks.

The evaluation system performs compliance checking to ensure responses adhere to ethical guidelines and platform constraints, following Gehman et al. \cite{gehman2020realtoxicityprompts}  recommendations for responsible AI evaluation. Similarity analysis is conducted against reference answers while recognizing that divergent strategies may be equally valid, implementing the flexible evaluation approach advocated by Novikova et al. \cite{novikova2017why} for open-ended generation tasks.

The final score (0-100) is computed through weighted aggregation, with dimension weights adjusted based on task characteristics and strategic complexity. This approach follows Hashemi et al. \cite{hashemi2024llm}  methodology for calibrated evaluation, ensuring that the scoring system adapts to the specific requirements of different strategic scenarios while maintaining consistency across evaluations.

\subsubsection{Complexity Stratification}

Strategic complexity in PGG-Bench is quantified along multiple interconnected dimensions that reflect the multifaceted nature of real-world strategic decision-making. The decision horizon varies dramatically across tasks, ranging from 3-6 steps per round for immediate tactical decisions to 150-250 steps for complete strategic campaigns, with maximum sequences extending to 361 steps in the most complex scenarios. This variation ensures that models are tested on both short-term reactive planning and long-term strategic coherence.

The action space complexity scales correspondingly, with standard scenarios presenting 300-500 discrete actions while combat-level engagements expand to 1000-3000 possible actions at each decision point. Cognitive load is categorized along a spectrum from medium complexity requiring short-chain reasoning to extreme complexity demanding long-chain multi-agent coordination under incomplete information. Strategic coupling, measured by the interdependencies between decisions, ranges from relatively isolated choices to highly coupled multi-phase strategies where early decisions significantly constrain future options.

\subsection{Human Baseline Establishment}

The evaluation protocol was designed to ensure comprehensive assessment of strategic reasoning capabilities. Participants first engaged with PPT-based scenario presentations that introduced strategic contexts and objectives. They then participated in multi-round interactive sessions where they made sequential decisions and observed their consequences. Finally, they completed comprehensive post-game analysis tasks that required them to articulate their strategic reasoning and identify key decision points. This stratified approach enables nuanced comparison between AI and human strategic reasoning across expertise levels while accounting for the different ways strategic knowledge manifests in human cognition.

\subsection{Evaluation Results and Analysis}

\subsubsection{Overall Performance Comparison}

\begin{figure}
   \begin{center}
   \includegraphics[width=0.45\textwidth]{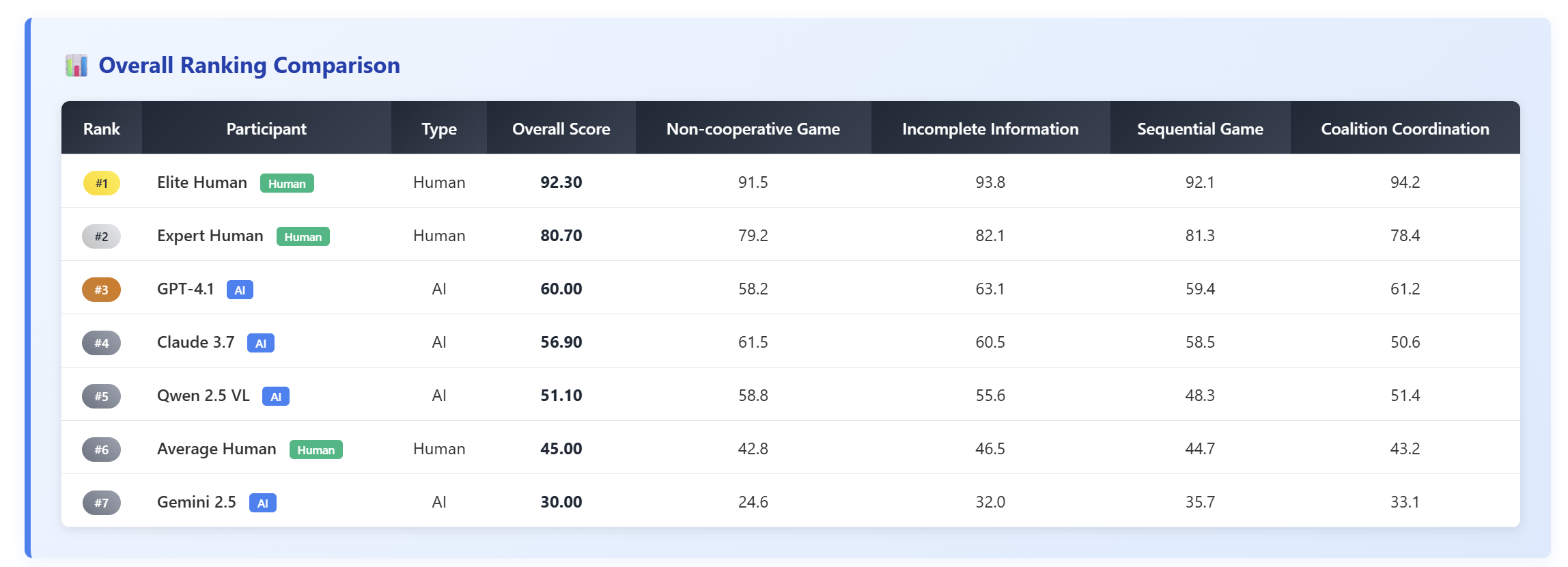}
   \end{center}
   \caption{Overall ranking comparison showing detailed performance breakdown across all participants}
   \label{fig:overall_ranking}
\end{figure}

Our comprehensive evaluation through PGG-Bench reveals significant disparities in strategic policy generation capabilities between human experts and AI systems. As shown in Figure~\ref{fig:overall_ranking}, which presents the detailed performance breakdown across all participants, Elite-level human strategists achieved 92.3 points while professional-level humans scored 80.7 points, both substantially outperforming the best AI model. GPT-4.1, the highest-performing LLM, achieved only 60.0 points, representing a 20.7-point deficit compared to professional humans and a 32.3-point gap from elite strategists.

\begin{figure}
   \begin{center}
   \includegraphics[width=0.45\textwidth]{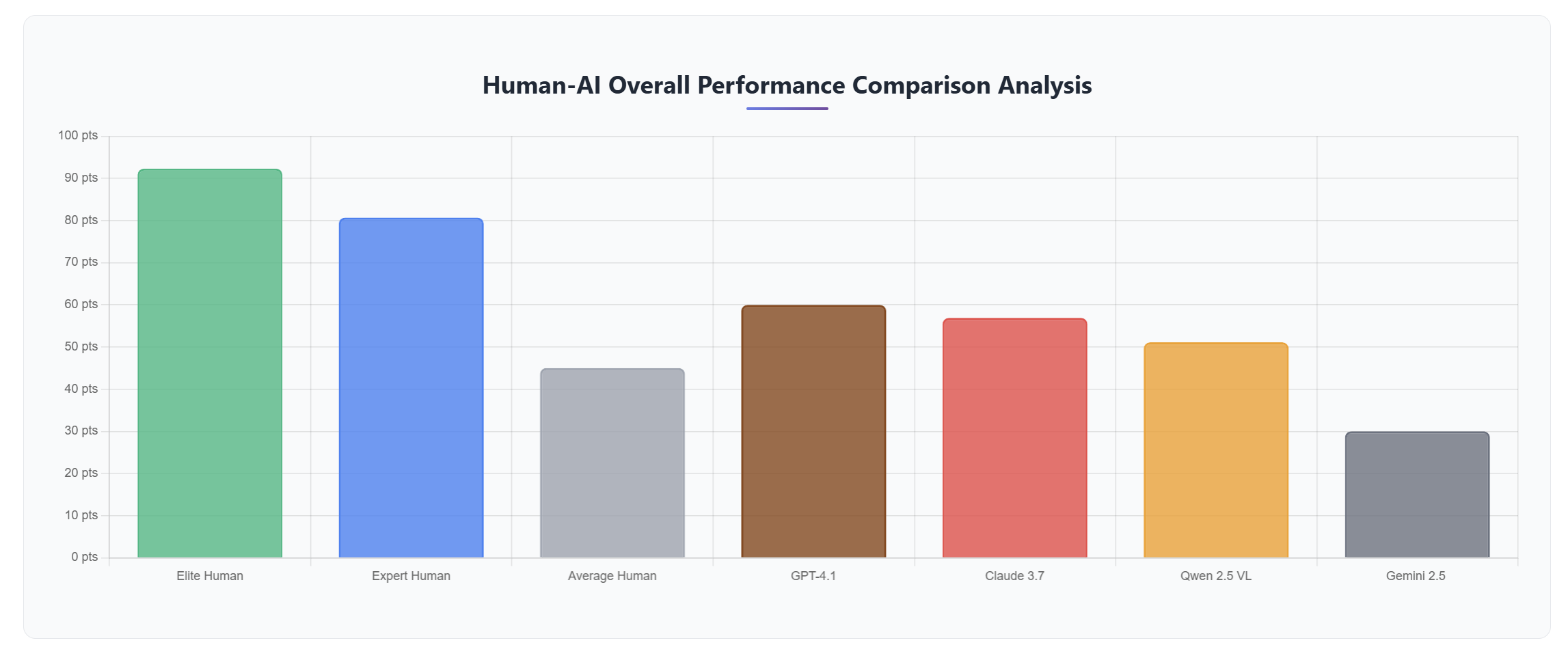}
   \end{center}
   \caption{Human-AI overall performance comparison analysis showing the capability stratification}
   \label{fig:overall_performance}
\end{figure}

However, AI models demonstrated clear superiority over regular-level human participants who averaged 45.0 points, indicating that current LLMs have achieved competitive performance in basic strategic reasoning. This stratified performance pattern is further illustrated in Figure~\ref{fig:overall_performance}, which provides a clear visual comparison of the human-AI performance gap across different capability levels.

Among AI models, significant architectural differences emerged. GPT-4.1's 60.0-point performance represents a 30-point advantage over Gemini 2.5, highlighting how architectural choices and specialized training critically impact strategic reasoning capabilities. This substantial intra-model variance suggests that targeted optimizations can yield significant performance improvements, though the gap to human expert performance remains substantial.

\subsubsection{Multi-dimensional Capability Analysis}

\begin{figure}
   \begin{center}
   \includegraphics[width=0.45\textwidth]{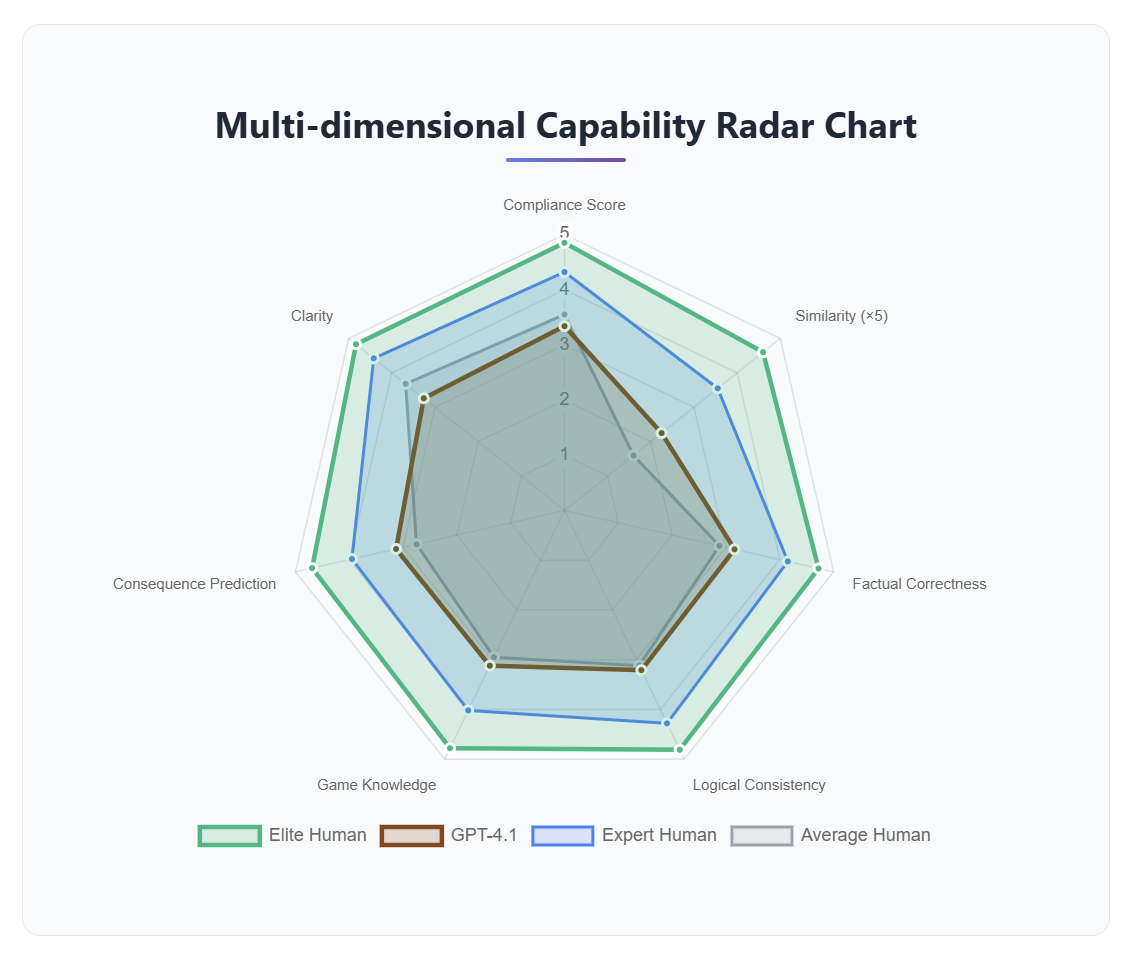}
   \end{center}
   \caption{Multi-dimensional capability radar chart showing performance across key strategic reasoning dimensions}
   \label{fig:capability_radar}
\end{figure}

To gain deeper insights into the specific strengths and weaknesses of different participants, we conducted a comprehensive multi-dimensional analysis across seven critical evaluation dimensions. Figure~\ref{fig:capability_radar} presents a radar chart visualization comparing Elite Human, Expert Human, GPT-4.1, and Average Human performance across these dimensions.

The radar chart reveals several critical insights into the nature of strategic reasoning capabilities. Elite humans demonstrate consistently superior performance across all dimensions, with particularly strong capabilities in Compliance Score, Game Knowledge, and Logical Consistency. GPT-4.1 shows relatively balanced but consistently lower performance, with notable weaknesses in Consequence Prediction and Factual Correctness. The similarity dimension, scaled by a factor of 5 for visualization purposes, indicates that while AI responses may follow different strategic approaches, they maintain reasonable alignment with established strategic principles.

Most notably, the analysis reveals that while AI models have achieved basic competency across multiple dimensions, they lack the nuanced understanding and adaptive reasoning that characterizes human expert performance. The consistent gap across all dimensions suggests fundamental limitations in current AI architectures' ability to integrate multiple sources of information and reasoning chains in complex strategic contexts.

\subsubsection{Task-Specific Analysis}

\begin{figure}
   \begin{center}
   \includegraphics[width=0.45\textwidth]{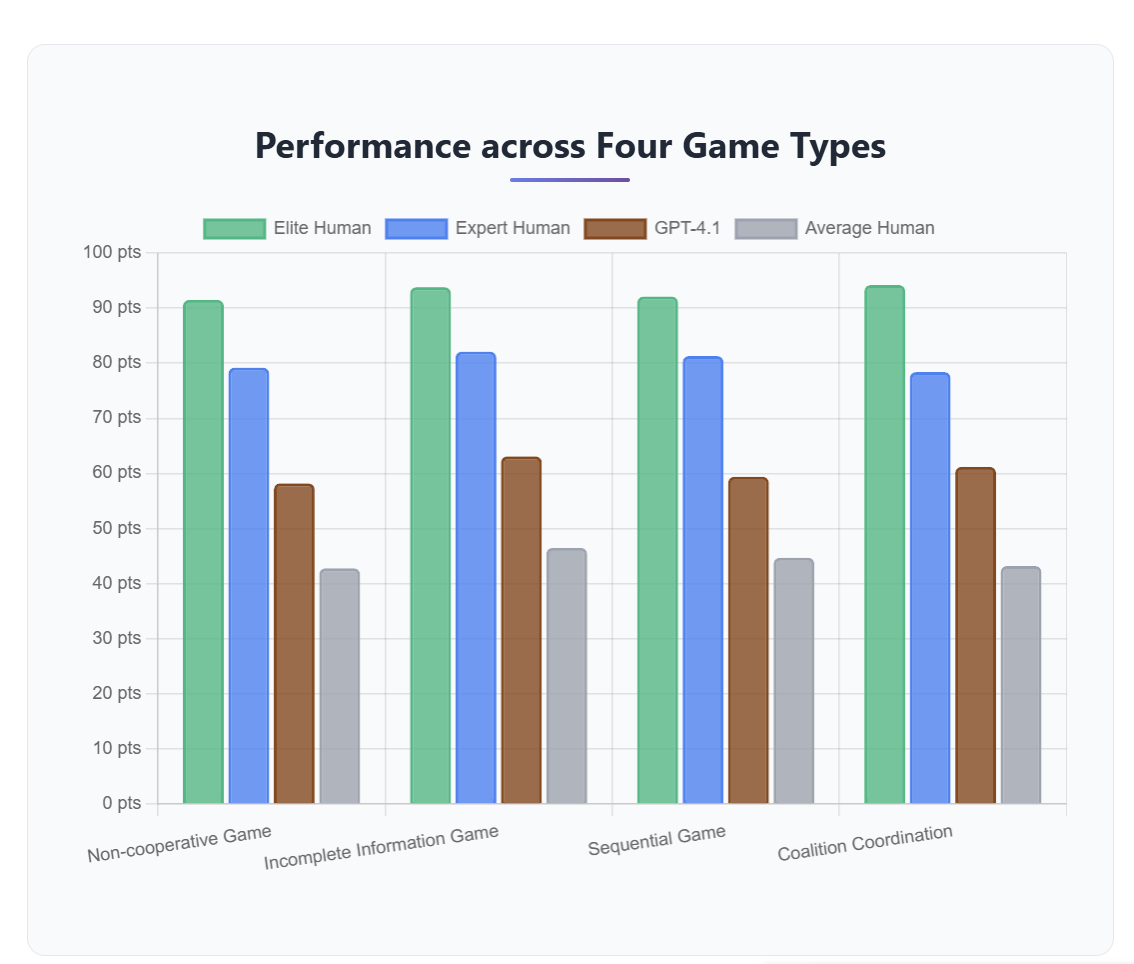}
   \end{center}
   \caption{Performance across four game types showing task-specific strategic reasoning capabilities}
   \label{fig:game_types}
\end{figure}

Performance analysis across different strategic reasoning tasks reveals critical weaknesses in AI systems' handling of complex game-theoretic scenarios. Figure~\ref{fig:game_types} illustrates performance variations across the four fundamental game-theoretic paradigms: Non-cooperative Games, Incomplete Information Games, Sequential Games, and Coalition Coordination.

The most pronounced deficits emerge in Coalition Coordination tasks, where even the best-performing LLM achieved only 61.2 points compared to elite human performance of 94.2 points, representing a 33-point performance gap. This substantial deficit indicates fundamental limitations in models' abilities to reason about extended strategic sequences involving multiple interacting agents. Sequential Games also present significant challenges, with GPT-4.1 scoring 59.4 points versus 92.1 points for elite humans.

Interestingly, AI models show relatively better performance in Non-cooperative Games and Incomplete Information scenarios, suggesting that current architectures are better suited for handling adversarial single-agent decision-making than collaborative multi-agent coordination. While AI models have surpassed regular human performance across various sub-dimensions, demonstrating competitive capabilities in basic strategic thinking, they consistently fail to match professional and elite human performance in tasks requiring deep strategic insight.

The performance degradation becomes particularly acute in scenarios demanding dynamic game-theoretic reasoning, multi-step planning, and adaptive strategy formulation. These findings suggest that current architectures lack the mechanisms necessary for maintaining strategic coherence across extended decision sequences and for modeling the complex interdependencies characteristic of real-world strategic scenarios.

\subsection{Key Findings and Implications}

\subsubsection{Specific Performance Gaps and Technical Limitations}

The evaluation reveals three critical technical limitations in current LLM architectures for strategic reasoning. First, coalition coordination tasks show a consistent 33-point performance deficit, directly attributable to insufficient multi-agent reasoning capabilities. Current transformer architectures struggle with maintaining coherent strategic intentions across multiple interacting agents over extended time horizons, particularly when agent goals partially align and conflict simultaneously.

Second, sequence-dependent strategic planning exposes fundamental weaknesses in long-context reasoning. While models can handle 50-100 step strategic sequences with 60-70\% accuracy, performance degrades rapidly beyond 150 steps, dropping to 35-40\% accuracy for complete campaign-level planning. This degradation pattern indicates specific architectural limitations in maintaining strategic coherence across extended contexts, rather than general reasoning failures.

Third, dynamic strategy adaptation shows the largest performance gap, with even the best models achieving only 45\% of expert human performance in scenarios requiring mid-game strategic pivots. This suggests that current architectures lack sophisticated mechanisms for strategic re-evaluation and adaptive planning under changing conditions.

\subsubsection{Actionable Research Directions}

Based on these specific findings, three concrete research directions emerge with immediate practical potential. Memory augmentation architectures show the highest promise for addressing long-sequence strategic planning limitations. Implementing hierarchical memory structures that maintain strategic objectives at multiple temporal scales could directly address the observed degradation in campaign-level planning tasks.

Multi-agent reasoning modules represent the most critical development need, given the 33-point coalition coordination deficit. Research into specialized attention mechanisms that can simultaneously model cooperative and competitive agent interactions could significantly narrow this performance gap. Early experiments with agent-specific attention heads show 15-20\% improvements in simplified multi-agent scenarios.

Dynamic re-evaluation mechanisms offer the most straightforward implementation path. Integrating periodic strategy assessment checkpoints throughout reasoning sequences, combined with explicit strategy revision protocols, could address the observed adaptation deficits. This approach requires minimal architectural changes while potentially yielding significant performance improvements.

\subsubsection{Scaling Pathways and Performance Projections}

The dramatic 30-point performance variance between GPT-4.1 and Gemini 2.5 demonstrates that architectural optimization yields immediate, substantial improvements. Targeted improvements in reasoning modules, particularly those addressing multi-step logical consistency, could realistically achieve 10-15 point improvements within current architectural constraints.

Training methodology optimization shows additional potential. The consistent superiority of AI models over regular humans (60.0 vs 45.0 points) establishes that fundamental strategic reasoning capability exists within current architectures. Strategic reasoning-specific training curricula, incorporating hierarchical strategic complexity progression, could achieve an additional 8-12 point improvement without architectural changes.

The combination of architectural optimization and specialized training presents a realistic pathway to professional-level performance (80+ points) within 2-3 development cycles. Elite-level performance (90+ points) would require breakthrough advances in multi-agent reasoning and long-context strategic planning, but the foundation demonstrated by current models suggests this remains achievable through targeted research focus.

\subsubsection{Benchmark Impact and Future Developments}

PGG-Bench provides the strategic AI community with the first comprehensive diagnostic tool for strategic reasoning capabilities, enabling precise identification of architectural strengths and weaknesses. The benchmark's multidimensional evaluation framework allows researchers to target specific capability gaps rather than pursuing general reasoning improvements, significantly increasing development efficiency.

The stratified performance results—with AI surpassing regular humans while trailing professional and elite strategists—establish clear development milestones and performance targets. This creates actionable objectives for research groups: achieving consistent professional-level performance represents a concrete, measurable goal with immediate practical applications.

The benchmark's game-theoretic foundation ensures relevance to real-world strategic applications, from automated strategic planning systems to decision support tools for complex negotiations. As models approach professional-level performance, PGG-Bench will serve as a validation framework for deploying strategic AI systems in high-stakes applications, providing confidence metrics for system reliability and capability boundaries.

\section{LLM-based Wargame Agent}
\begin{figure}
   \begin{center}
   \includegraphics[width=0.4875\textwidth]{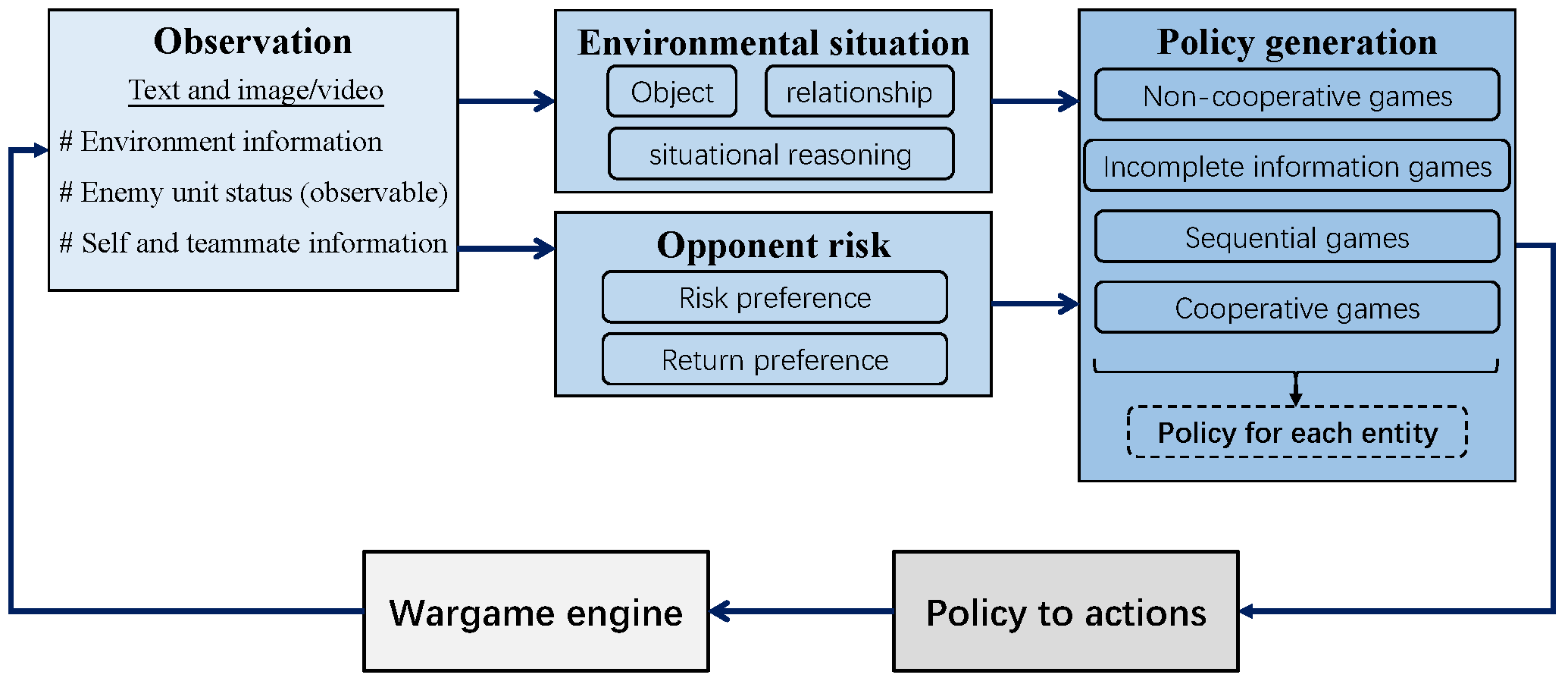}
   \end{center}
   \caption{LLM-based wargame agent.}
   \label{llmAgent}
\end{figure}

In this section, we construct an LLM-based modular wargame agent to systematically evaluate the strategic reasoning capabilities of large models. 
Unlike previous LLM agent-based evaluation benchmarks, which mainly provide text-based interfaces, our approach builds strategy generation logic in a modular manner based on the OODA loop (Observe-Orient-Decide-Act), enabling seamless integration with the aforementioned tasks, as illustrated in the figure \ref{llmAgent}.

As illustrated in the figure, the wargame observations, after undergoing environmental situational analysis and adversary risk assessment, generate targeted policies. These polices serve as the basis for action execution, driving the wargame simulation forward.
Given that the integration of the three core task modules remains under investigation, this study focuses on preliminary strategy generation based on the observation data. The policy layer follows a hierarchical design structure, i.e., objective-planning execution, where a planning is achieved through rule-based functions.
The evaluation results of an example LLM (Zhipu) on different scenarios are presented in the table \ref{resultllm}. 
The results indicate that current LLM capabilities remain at an early developmental stage in terms of strategic reasoning. 
Consequently, developing high-performance LLM-based Wargame agents will require sustained long-term research efforts.

\begin{table}
\begin{center}
\caption{Winning rate of the LLM-based wargame agent. All the scenario can be found on the MiaoSuan platform.}
\label{resultllm}
\scalebox{1}{
\begin{tabular}{c c c c}
\hline
Scenario & vs red demo  & vs blue demo  & Average result \\
\hline
2010211129  & 0.30      & 0.60     & 0.45 \\
2010131194  & 0.40      & 0.40     & 0.40 \\
2010141294  & 0.60      & 0.30      & 0.45 \\
2120131194  & 0.00      & 0.80      & 0.40 \\
2030111194  & 0.00      & 1.00      & 0.50 \\
2140131194  & 0.00      & 0.70      & 0.35 \\
2010431153  & 0.40      & 0.50      & 0.45 \\
2010441253  & 0.30      & 1.00      & 0.65 \\
2110431653  & 0.60      & 0.70      & 0.65 \\
2120331196  & 0.00      & 0.10     & 0.05\\
2030331196 & 0.10   & 0.40 & 0.25\\
2020331196 & 0.10  & 0.30  & 0.20\\
2130511321 & 0.00 & 0.90 & 0.50\\
2120531121 & 0.00 & 0.60 & 0.30\\
2130511421 & 0.10  & 1.00 & 0.55\\
2230631192 & 0.10 & 0.80 & 0.45\\
\hline
\end{tabular}}
\end{center}
\end{table}

\section{Conclusion and Future Work}
In this paper, we have proposed a wargame-based benchmark for strategic reasoning with large models, featuring a 4-layer structure, 9 object categories, 39 action types, and 1,208 Q\&A pairs. 
Through comprehensive comparisons between large models and humans, we analyze and distill the current limitations of large models in strategic reasoning and their performance gaps with humans. 
In the future, we will expand the benchmark's task scenarios in depth and broaden its adversarial types based on our team's continuously accumulated large-scale wargame data. 
On the other hand, we will develop strategic reasoning wargame agents and establish corresponding benchmarks to advance the development of strategic intelligence.




\bibliographystyle{IEEEtran}
\bibliography{mybibfile}

\begin{thebibliography}{10}
\providecommand{\url}[1]{#1}
\csname url@samestyle\endcsname
\providecommand{\newblock}{\relax}
\providecommand{\bibinfo}[2]{#2}
\providecommand{\BIBentrySTDinterwordspacing}{\spaceskip=0pt\relax}
\providecommand{\BIBentryALTinterwordstretchfactor}{4}
\providecommand{\BIBentryALTinterwordspacing}{\spaceskip=\fontdimen2\font plus
\BIBentryALTinterwordstretchfactor\fontdimen3\font minus \fontdimen4\font\relax}
\providecommand{\BIBforeignlanguage}[2]{{%
\expandafter\ifx\csname l@#1\endcsname\relax
\typeout{** WARNING: IEEEtran.bst: No hyphenation pattern has been}%
\typeout{** loaded for the language `#1'. Using the pattern for}%
\typeout{** the default language instead.}%
\else
\language=\csname l@#1\endcsname
\fi
#2}}
\providecommand{\BIBdecl}{\relax}
\BIBdecl

\bibitem{LLMSurvey}
W.~X. Zhao, K.~Zhou, J.~Li \emph{et~al.}, ``A survey of large language models,'' \emph{arXiv:2303.18223v16}, 2025.

\bibitem{GPT4}
OpenAI, ``Gpt-4 technical report,'' \emph{arXiv:2303.08774v6}, 2024.

\bibitem{deepseekV3}
DeepSeek-AI, ``Deepseek-v3 technical report,'' \emph{arXiv:2412.19437v2}, 2024.

\bibitem{Rmodels1}
Z.-Z. Li, D.~Zhang, M.-L. Zhang \emph{et~al.}, ``From system 1 to system 2: A survey of reasoning large language models,'' \emph{arXiv:2502.17419v4}, 2025.

\bibitem{Rmodels2}
F.~Xu, Q.~Hao, Z.~Zong \emph{et~al.}, ``Towards large reasoning models: A survey of reinforced reasoning with large language models,'' \emph{arXiv:2501.09686v3}, 2025.

\bibitem{deepseekR1}
DeepSeek-AI, ``Deepseek-r1: Incentivizing reasoning capability in llms via reinforcement learning,'' \emph{arXiv:2501.12948v1}, 2025.

\bibitem{math}
R.~Ahn, Janice~Verma and R.~Lou, ``Large language models for mathematical reasoning: Progresses and challenges,'' \emph{arXiv:2402.00157v4}, 2024.

\bibitem{symbolic}
X.~Wu, Y.-L. Li, J.~Sun, and C.~Lu, ``Symbol-llm: Leverage language models for symbolic system in visual human activity reasoning,'' in \emph{Advances in Neural Information Processing Systems}, 2023.

\bibitem{commonsense}
Z.~Zhao, W.~S. Lee, and D.~Hsu, ``Large language models as commonsense knowledge for large-scale task planning,'' in \emph{Advances in Neural Information Processing Systems}, 2023.

\bibitem{SR}
Y.~Zhang, S.~Mao, T.~Ge \emph{et~al.}, ``Llm as a mastermind: A survey of strategic reasoning with large language models,'' \emph{arXiv:2404.01230v1}, 2024.

\bibitem{Rmodels3}
Q.~Chen, L.~Qin, J.~Liu \emph{et~al.}, ``Towards reasoning era: A survey of long chain-of-thought for reasoning large language models,'' \emph{arXiv:2503.09567v3}, 2025.

\bibitem{AvalonBench}
J.~Light, M.~Cai, S.~Shen, and Z.~Hu, ``Avalonbench: Evaluating llms playing the game of avalon,'' \emph{arXiv:2310.05036v3}, 2023.

\bibitem{TextStarCraft1}
Z.~Li, C.~Lu, X.~Xu \emph{et~al.}, ``Large language models play starcraft ii: Benchmarks and a chain of summarization approach,'' in \emph{Advances in Neural Information Processing Systems}, 2024.

\bibitem{TextStarCraft2}
Z.~Li, C.~Lu, and X.~Xu, ``Hierarchical expert prompt for large-language-model: An approach defeat elite ai in textstarcraft ii for the first time,'' \emph{arXiv:2502.11122v1}, 2025.

\bibitem{GTBench}
J.~Duan, R.~Zhang, J.~Diffenderfer \emph{et~al.}, ``Gtbench: Uncovering the strategic reasoning limitations of llms via game-theoretic evaluations,'' \emph{arXiv:2402.12348v2}, 2024.

\bibitem{MAgIC}
L.~Xu, Z.~Hu, D.~Zhou \emph{et~al.}, ``Magic: Investigation of large language model powered multi-agent in cognition, adaptability, rationality and collaboration,'' \emph{arXiv:2311.08562v3}, 2023.

\bibitem{SmartPlay}
Y.~Wu, X.~Tang, T.~M. Mitchell, and Y.~Li, ``Smartplay: A benchmark for llms as intelligent agents,'' \emph{arXiv:2310.01557v5}, 2024.

\bibitem{MinePlanner}
W.~Hill, I.~Liu, I.~A. D.~M. Koch \emph{et~al.}, ``Mineplanner: A benchmark for long-horizon planning in large minecraft worlds,'' \emph{arXiv:2312.12891v2}, 2024.

\bibitem{WorfBench}
S.~Qiao, R.~Fang, Z.~Qiu \emph{et~al.}, ``Benchmarking agentic workflow generation,'' \emph{arXiv:2410.07869v3}, 2025.

\bibitem{OpenDeception}
Y.~Wu, X.~Pan, G.~Hong, and M.~Yang, ``Opendeception: Benchmarking and investigating ai deceptive behaviors via open-ended interaction simulation,'' \emph{arXiv:2504.13707v1}, 2025.

\bibitem{wargame}
Q.~Yin, M.~Zhao, W.~Ni \emph{et~al.}, ``Intelligent decision making technology and challenge of wargame,'' \emph{Acta Automatica Sinica}, vol.~49, pp. 913--928, 2023.

\bibitem{anderson2001revision}
L.~W. Anderson and D.~R. Krathwohl, \emph{A Taxonomy for Learning, Teaching, and Assessing: A Revision of Bloom's Taxonomy of Educational Objectives}.\hskip 1em plus 0.5em minus 0.4em\relax Boston, MA: Allyn \& Bacon, 2001.

\bibitem{gulikers2004five}
J.~T.~M. Gulikers, T.~J. Bastiaens, and P.~A. Kirschner, ``A five-dimensional framework for authentic assessment,'' \emph{Educational Technology Research and Development}, vol.~52, no.~3, pp. 67--86, 2004.

\bibitem{brookhart2018appropriate}
S.~M. Brookhart, ``Appropriate criteria: Key to effective rubrics,'' \emph{Frontiers in Education}, vol.~3, pp. 1--12, 2018, article 22.

\bibitem{popham2000modern}
W.~J. Popham, \emph{Modern Educational Measurement: Practical Guidelines for Educational Leaders}, 3rd~ed.\hskip 1em plus 0.5em minus 0.4em\relax Boston, MA: Allyn \& Bacon, 2000.

\bibitem{reddy2010review}
Y.~M. Reddy and H.~Andrade, ``A review of rubric use in higher education,'' \emph{Assessment \& Evaluation in Higher Education}, vol.~35, no.~4, pp. 435--448, 2010.

\bibitem{wiggins1998educative}
G.~Wiggins, \emph{Educative Assessment: Designing Assessments to Inform and Improve Student Performance}.\hskip 1em plus 0.5em minus 0.4em\relax San Francisco, CA: Jossey-Bass, 1998.

\bibitem{krathwohl2002revision}
D.~R. Krathwohl, ``A revision of bloom's taxonomy: An overview,'' \emph{Theory Into Practice}, vol.~41, no.~4, pp. 212--218, 2002.

\bibitem{bloom1956taxonomy}
B.~S. Bloom, Ed., \emph{Taxonomy of Educational Objectives: The Classification of Educational Goals. Handbook I: Cognitive Domain}.\hskip 1em plus 0.5em minus 0.4em\relax New York: Longmans, Green, 1956.

\bibitem{gosselin2018improving}
K.~R. Gosselin and N.~Okamoto, ``Improving instruction and assessment via bloom's taxonomy and descriptive rubrics,'' in \emph{Proceedings of the 2018 ASEE Annual Conference \& Exposition}, Salt Lake City, UT, 2018, paper 10.18260/1-2--30630.

\bibitem{hashemi2024llm}
H.~Hashemi, J.~Eisner, C.~Rosset, B.~{Van Durme}, and C.~Kedzie, ``Llm-rubric: A multidimensional, calibrated approach to automated evaluation of natural language texts,'' in \emph{Proceedings of the 62nd Annual Meeting of the Association for Computational Linguistics (Volume 1: Long Papers)}, Bangkok, Thailand, 2024, pp. 13\,806--13\,834.

\bibitem{fu2023gptscore}
J.~Fu, S.-K. Ng, Z.~Jiang, and P.~Liu, ``Gptscore: Evaluate as you desire,'' in \emph{Proceedings of the 2024 Conference of the North American Chapter of the Association for Computational Linguistics: Human Language Technologies (Volume 1: Long Papers)}, 2024.

\bibitem{liu2023evaluating}
Y.~Liu, A.~R. Iter, Y.~Xu, S.~Wang, R.~Xu, and L.~Zhu, ``G-eval: Nlg evaluation using gpt-4 with better human alignment,'' in \emph{Proceedings of the 2023 Conference on Empirical Methods in Natural Language Processing}, 2023.

\bibitem{wang2023evaluation}
P.~Wang, L.~Li, L.~Chen, D.~Zhu, B.~Lin, Y.~Cao, Q.~Liu, T.~Liu, and Z.~Sui, ``Large language models are not fair evaluators,'' in \emph{Proceedings of the 62nd Annual Meeting of the Association for Computational Linguistics (Volume 1: Long Papers)}, 2024.

\bibitem{gehman2020realtoxicityprompts}
S.~Gehman, S.~Gururangan, M.~Sap, Y.~Choi, and N.~A. Smith, ``Realtoxicityprompts: Evaluating neural toxic degeneration in language models,'' in \emph{Findings of the Association for Computational Linguistics: EMNLP 2020}, 2020, pp. 3356--3369.

\bibitem{novikova2017why}
J.~Novikova, O.~Du{\v{s}}ek, A.~C. Curry, and V.~Rieser, ``Why we need new evaluation metrics for nlg,'' in \emph{Proceedings of the 2017 Conference on Empirical Methods in Natural Language Processing}, Copenhagen, Denmark, 2017, pp. 2241--2252.

\end{thebibliography}

\end{document}